\documentclass[11pt]{article}
\usepackage[linesnumbered,ruled,vlined]{algorithm2e}
\usepackage{amsfonts}
\usepackage{amsmath}
\usepackage{arydshln}
\usepackage{authblk}
\usepackage{hhline}
\usepackage{bm}
\usepackage{booktabs}
\usepackage[margin=1in]{geometry}
\usepackage{graphicx}
\usepackage{mathtools}
\usepackage{multirow}
\usepackage{xcolor}
\usepackage[colorlinks=true,citecolor=blue]{hyperref}
\graphicspath{{figs/}}

\title{Reliable extrapolation of deep neural operators informed by physics or sparse observations}
\author[1]{Min Zhu}
\author[2]{Handi Zhang}
\author[3]{Anran Jiao}
\author[4,5]{George Em Karniadakis}
\author[1,*]{Lu Lu}
\affil[1]{Department of Chemical and Biomolecular Engineering, University of Pennsylvania, Philadelphia, PA 19104, USA}
\affil[2]{Graduate Group in Applied Mathematics and Computational Science, University of Pennsylvania, Philadelphia, PA 19104, USA}
\affil[3]{Department of Computer and Information Science, University of Pennsylvania, Philadelphia, PA 19104, USA}
\affil[4]{Division of Applied Mathematics, Brown University, Providence, RI 02912, USA}
\affil[5]{School of Engineering, Brown University, Providence, RI 02912, USA}
\affil[*]{Corresponding author. Email: lulu1@seas.upenn.edu}
\date{}

\begin{document}

\maketitle

\begin{abstract}
Deep neural operators can learn nonlinear mappings between infinite-dimensional function spaces via deep neural networks. As promising surrogate solvers of partial differential equations (PDEs) for real-time prediction, deep neural operators such as deep operator networks (DeepONets) provide a new simulation paradigm in science and engineering. Pure data-driven neural operators and deep learning models, in general, are usually limited to interpolation scenarios, where new 
predictions utilize inputs within the support of the training set. However, in the inference stage of real-world applications, the input may lie outside the support, i.e., extrapolation is required, which may result to large errors and unavoidable failure of deep learning models. Here, we address this challenge of extrapolation for deep neural operators. First, we systematically investigate the extrapolation behavior of DeepONets by quantifying the extrapolation complexity via the 2-Wasserstein distance between two function spaces and propose a new behavior of bias-variance trade-off for extrapolation with respect to model capacity. Subsequently, we develop a complete workflow, including extrapolation determination, and we propose five reliable learning methods that guarantee a safe prediction under extrapolation by requiring additional information---the governing PDEs of the system or sparse new observations. The proposed methods are based on either fine-tuning a pre-trained DeepONet or multifidelity learning. We demonstrate the effectiveness of the proposed framework for various types of parametric PDEs. Our systematic comparisons provide practical guidelines for selecting a proper extrapolation method depending on the available information, desired accuracy, and required inference speed.
\end{abstract}


\section{Introduction}
\label{sec:intro}

The universal approximation theorem of neural networks (NNs) for functions~\cite{hornik1989multilayer} has provided a rigorous foundation of deep learning. As an increasingly popular alternative to traditional numerical methods such as finite difference and finite element methods, neural networks have been applied in solving partial differential equations (PDEs) in the field of scientific machine learning (SciML)~\cite{baker2019workshop,karniadakis2021physics}. Physics-informed neural networks (PINNs)~\cite{RAISSI2019686,lu2021deepxde} have provided a new paradigm for solving forward as well as inverse problems governed by PDEs by embedding the PDE loss into the loss function of neural networks.
% PINNs have been utilized for diverse applications~\cite{chen2020physics,yazdani2020systems,daneker2022systems, mao2020physics} and have been extended to solve integro-differential equations~\cite{lu2021deepxde}, fractional PDEs~\cite{fPINNs}, and stochastic PDEs~\cite{zhang2019quantifying}. Variants of PINNs have also been developed to improve the accuracy and efficiency, such as gradient-enhanced PINNs (gPINNs)~\cite{YU2022114823}, PINNs with hard constraints (hPINNs)~\cite{hPINN}, \textcolor{red}{variational PINNs~\cite{kharazmi2019variational,kharazmi2021hp}} residual-based adaptive sampling~\cite{lu2021deepxde,wu2022comprehensive}, \textcolor{red}{parallel PINNs~\cite{SHUKLA2021110683}, and non-Newtonian PINNs~\cite{D1SM01298C}}.

Neural networks are universal approximators of not only functions but also nonlinear operators, i.e., mappings between infinite-dimensional function spaces~\cite{chen1995universal,deeponetNatureML,deng2022approximation}. Hence, NNs can approximate the operators of PDEs, and once the network is trained, it only requires a forward pass to obtain the PDE solution for a new condition. The deep operator network (DeepONet)~\cite{deeponetNatureML}, the first neural operator, has  demonstrated good performance for building surrogate models for many types of PDEs. For example, DeepONet has been applied in multiscale bubble dynamics~\cite{lin2021operator,lin_maxey_li_karniadakis_2021}, brittle fracture analysis~\cite{GOSWAMI2022114587},  instabilities in boundary layers~\cite{di2021deeponet}, solar-thermal systems forecasting~\cite{osti_1839596}, electroconvection~\cite{cai2021deepm}, hypersonics with chemical reactions~\cite{MAO2021110698}, and fast multiscale modeling~\cite{yin2022interfacing}. In addition, several extensions of DeepONet have been proposed in recent studies, including DeepONet for multiple-input operators (MIONet)~\cite{jin2022mionet}, DeepONet with proper orthogonal decomposition (POD-DeepONet)~\cite{lu2022comprehensive}, physics-informed DeepONet~\cite{wang2021learning,GOSWAMI2022114587}, multifidelity DeepONet~\cite{lu2022multifidelity,howard2022multifidelity,de2022bi}, DeepM\&Mnet for multiphysics problems~\cite{cai2021deepm,MAO2021110698}, multiscale DeepONet~\cite{liu2021multiscale}, and DeepONet with uncertainty quantification~\cite{lin2021accelerated,psaros2022uncertainty,uqdeeponet,moya2022deeponet}.

\begin{figure}[htbp]
  \centering
  \includegraphics[width=\textwidth]{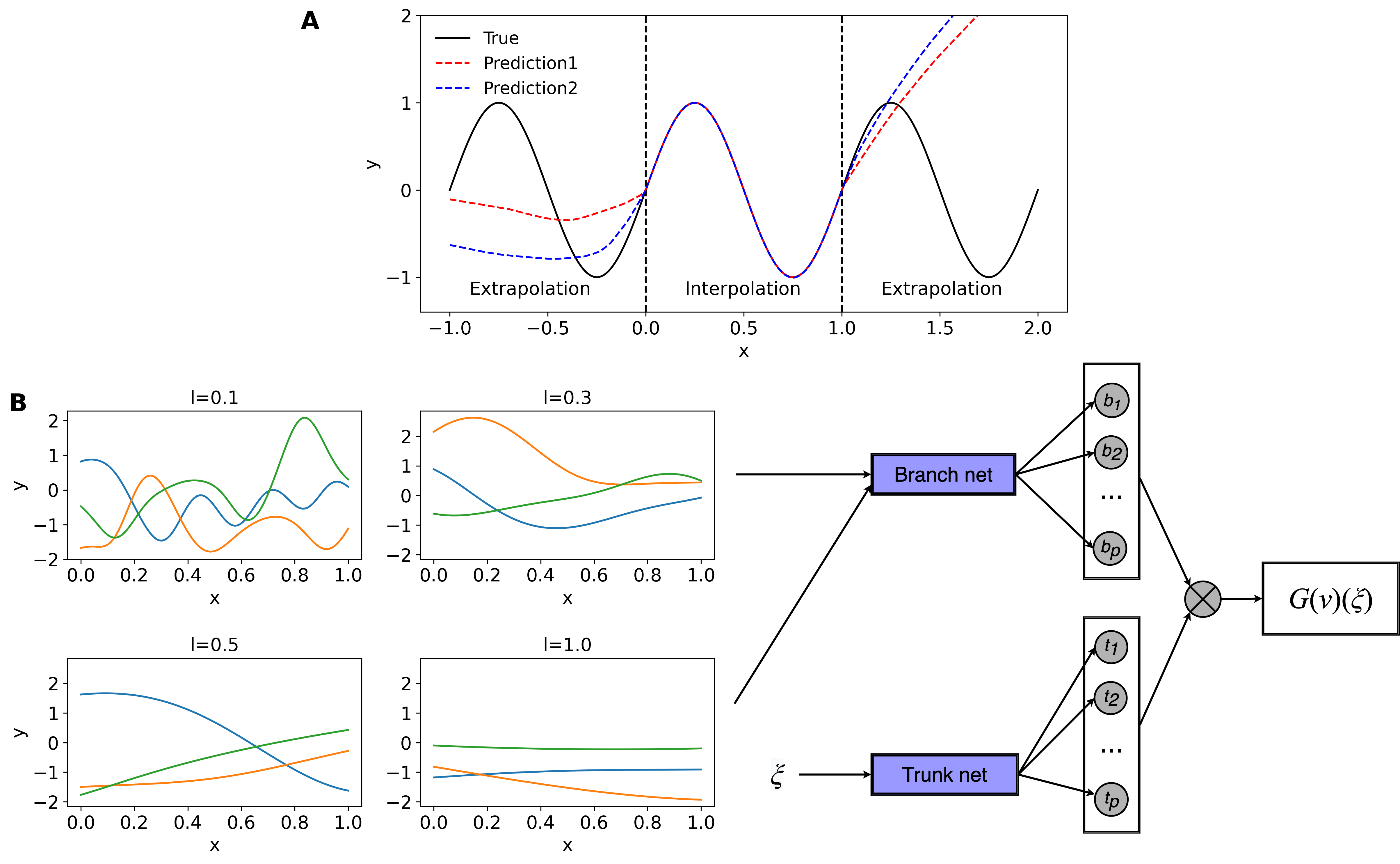}
  \caption{\textbf{Examples of NNs and DeepONets for interpolation and extrapolation.} (\textbf{A}) Two independent NNs are trained to learn $y=\sin(2\pi x)$ using the data in $[0, 1]$. (\textbf{B}) Functions randomly sampled from GRF spaces of different correlation lengths $l$ are taken as the input functions for the branch net of DeepONet.}
  \label{fig:fig_1}
\end{figure}

Despite the aforementioned success, neural networks are usually limited to solving interpolation problems, i.e., inference is accurate only for inputs within the support of the training set, while NNs would fail if the new input is outside the support of the training set (i.e., extrapolation)~\cite{158898,xu2020neural}. We provide an illustrative example in Fig.~\ref{fig:fig_1}A, where we trained two fully-connected neural networks (three hidden layers, 64 neurons per layer, and ReLU activation function) to learn the ground-truth function $y=\sin(2\pi x)$. The training data is 100 equispaced points in the domain of $[0, 1]$. After training, the two NNs perform well in the interpolation region. However, the two networks have very large prediction error in the extrapolation region ($x < 0$ or $x > 1$).

This difficulty of extrapolation has also been observed in DeepONets~\cite{deeponetNatureML,lin2021operator,MAO2021110698,kissas2022learning,yin2022interfacing,ppnn}. When we use DeepONets to learn an operator $\mathcal{G}: v(x) \mapsto u(\xi)$, we usually generate a training dataset by randomly sampling input functions $v$ from a space of mean-zero Gaussian random field (GRF; or Gaussian process (GP)) with a predefined covariance kernel $k_l(x_1, x_2)$:
$$v(x) \sim \mathcal{GP}(0, k_l(x_1, x_2)).$$
A typical choice of the kernel is the radial-basis function kernel (or squared-exponential kernel, Gaussian kernel)
\begin{equation} \label{eq:rbf}
    k_l(x_1, x_2) = \exp \left(-\frac{\|x_1 - x_2 \|^2}{2l^2} \right),
\end{equation}
where $l>0$ is the correlation length. The value of $l$ determines the smoothness of sampled functions, with a larger $l$ leading to smoother functions. Several functions randomly sampled from the spaces of GRFs with different $l$ are shown in Fig.~\ref{fig:fig_1}B.

In real applications, there is no guarantee that in the inference stage, a new input is always in the interpolation region, and extrapolation would lead to large errors and possible failure of NNs. In this study, our first goal is to understand the extrapolation of DeepONets. We quantitatively measure the extrapolation complexity via the 2-Wasserstein distance between two function spaces. Subsequently, we systematically study how the extrapolation error of DeepONet changes with respect to multiple factors, including model capacity (e.g., network size and training iterations), training dataset size, and activation functions.

In the second part of this work, we address the extrapolation issue for DeepONets. Given a training dataset, in general, it is almost impossible to have an accurate prediction for a new input outside the support of training dataset. In order to reduce the extrapolation error, we assume that we have one of the following extra information: (1) we know the governing PDE of the system, or (2) we have extra sparse observations at some locations $\xi$ for the output function $u$. Based on the type of information, we propose several methods to solve the extrapolation problem. Specifically, we first train a DeepONet by using the training dataset. Next, for the first case with physics, we fine-tune the pre-trained DeepONet with the PDE loss as done in PINNs. For the second case with new observations, we propose to either fine-tune the pre-trained DeepONet with the new data, or train another machine learning model (neural networks or Gaussian process regression) using multifidelity learning~\cite{kennedy2000predicting,sobester2008engineering,meng2020composite,lu2020extraction,lu2022machine}, where the prediction from the pre-trained DeepONet is of low-fidelity while the new data is of high-fidelity.

The idea of using new observations to fine-tune a pre-trained DeepONet for extrapolation has been used in Refs.~\cite{lin2021operator,MAO2021110698}. However, as we show in our numerical experiments, their fine-tuning approach is not always stable and accurate, so here we propose a better approach for fine-tuning. Moreover, fine-tuning a pre-trained network with new data is conceptually related to transfer learning (TL)~\cite{pan2009survey,zhuang2020comprehensive,deng2021deep} and few-shot learning (FSL)~\cite{lu2020learning,wang2020generalizing}. In Ref.~\cite{goswami2022deep}, a transfer learning technique was developed for DeepONets to solve the same PDEs but on different domains. Although we only consider DeepONets in this study, the proposed methods can directly be applied to other deep neural operators, such as Fourier neural operators~\cite{li2020fourier,lu2022comprehensive}, graph kernel networks~\cite{li2020neural}, nonlocal kernel networks~\cite{you2022nonlocal}, and others~\cite{trask2019gmls,kissas2022learning,patel2021physics}.

The paper is organized as follows. In Section~\ref{sec:extp}, after briefly introducing the architecture of DeepONet, we first introduce a definition of the extrapolation complexity and subsequently empirically investigate the extrapolation error with respect to various factors. In Section~\ref{sec:methods}, we provide a general workflow for extrapolation and propose several methods to improve the DeepONet performance for extrapolation, including fine-tuning with physics, fine-tuning with sparse observations, and multifidelity learning. Then in Section~\ref{sec:results}, we compare the performance of different methods in six numerical examples. Finally, we conclude the paper and discuss some future directions in  Section~\ref{sec:conclusion}.

\section{Extrapolation of deep neural operators}
\label{sec:extp}

We briefly explain the architecture of the deep operator network (DeepONet) and then focus on how extrapolation errors vary with respect to different factors, including extrapolation complexity, training phase, training dataset size, and network architecture (e.g., network size and activation function).

\subsection{Operator learning via DeepONet}
\label{sec:deeponet}

To define the setup of operator learning, we consider a function space $\mathcal{V}$ of function $v$ defined on the domain $D \subset \mathbb{R}^d$:
$$
v: D \ni x \mapsto v(x) \in \mathbb{R},
$$
and another function space $\mathcal{U}$ of function $u$ defined on the domain $D'\subset \mathbb{R}^{d'} $. 
$$
u: D' \ni \xi \mapsto u(\xi) \in \mathbb{R}.
$$
Let $\mathcal{G}$ be an operator that maps $\mathcal{V}$ to $\mathcal{U}$:
$$
\mathcal{G}: \mathcal{V} \to \mathcal{U}, \quad v \mapsto u.
$$

DeepONet~\cite{deeponetNatureML} was developed to learn the operator $\mathcal{G}$ based on the universal approximation theorem of neural networks for operators~\cite{chen1995universal}. A DeepONet consists of two sub-networks: a trunk network and a branch network (Fig.~\ref{fig:fig_1}B). For $m$ scattered locations $\{x_1, x_2, \dots, x_m\}$ in $D$, the branch network takes the function evaluations $[v(x_1), v(x_2), \dots, v(x_m)]$ as the input, and the output of the branch network is $[b_1(v), b_2(v), \dots, b_p(v)]$, where $p$ is the number of neurons. The trunk network takes $\xi$ as the input and the outputs $[t_1(\xi), t_2(\xi), \dots, t_p(\xi)]$. Then, by taking the inner product of trunk and branch outputs, the output of DeepONet is:
$$
\mathcal{G}(v)(\xi) = \sum_{k=1}^p b_k(v)t_k(\xi) + b_0,
$$
where $b_0 \in \mathbb{R}$ is a bias.

\subsection{Experiment setup}
\label{subsec:2.2}
We consider two examples to demonstrate the extrapolation of DeepONets.

\paragraph{Antiderivative operator.}
The first one is an ordinary differential equation (ODE) defined by
\begin{equation*}
    \frac{du(x)}{dx} = v(x), \quad x\in [0,1],
\end{equation*}
with an initial condition $u(0)=0$. We use DeepONet to learn the solution operator, i.e., an antiderivative operator
$$\mathcal{G}: v(x) \mapsto u(x)=\int_0^x v(\tau)d\tau.$$

\paragraph{Diffusion-reaction equation.}
% TODO: k=0.1/0.05 (Min)
The second example is a diffusion-reaction equation defined by
\begin{equation*}
    \frac{\partial u}{\partial t} = D \frac{\partial^2 u}{\partial x^2} + ku^2 + v(x), \quad x \in [0,1], t \in [0,1],
\end{equation*}
with zero initial and boundary conditions. In this example, $k$ and $D$ are set at 0.01. DeepONet is trained to learn the mapping from the source term $v(x)$ to the solution $u(x, t)$:
$$\mathcal{G}: v(x) \mapsto u(x,t).$$

\paragraph{Training and test datasets.}
The input functions $v$ are sampled from a mean-zero Gaussian random field (GRF) $v \sim \mathcal{GP}(0, k_l(x_1,x_2))$ with the radial-basis function (RBF) kernel of Eq.~\eqref{eq:rbf}. However, the training and test datasets use different values of $l$. The reference solution $u(x)$ of the ODE is obtained by Runge-Kutta(4, 5), and the reference solution $u(x,t)$ of the PDE is obtained by a second-order finite difference method with a mesh size of $101\times 101$.

\subsection{Interpolation and extrapolation regions}

The input functions for training and testing are generated from GRFs, with a larger $l$ leading to smoother functions (Fig.~\ref{fig:fig_1}B). Hence, if the correlation length for training ($l_{\text{train}}$) is different from the correlation length for testing ($l_{\text{test}}$), then it is extrapolation (Ex.). The level of extrapolation can be represented by the difference between $l_{\text{train}}$ and $l_{\text{test}}$.

\begin{figure}[htbp]
    \centering
    \includegraphics[width=\textwidth]{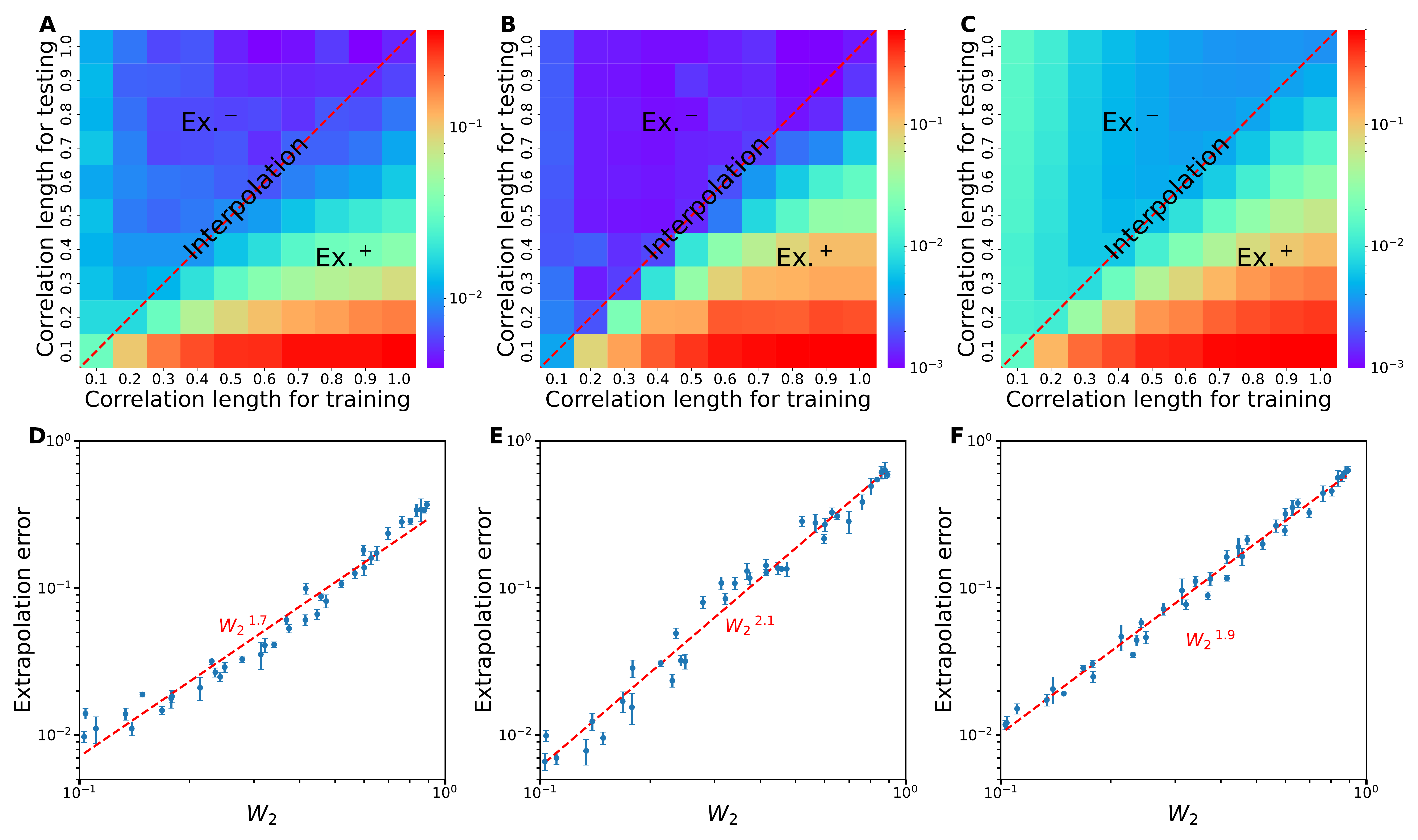}
    \caption{\textbf{$L^2$ relative errors of DeepONets trained and tested on datasets with different correlation lengths.} (\textbf{A} and \textbf{D}) ODE problem. (\textbf{B} and \textbf{E}) Diffusion-reaction equation with $k=0.01$. (\textbf{C} and \textbf{F}) Diffusion-reaction equation with $k=0.5$. (A--C) The testing error for different pairs of training and testing functions. (D--F) The Ex.\textsuperscript{+} error grows with a polynomial rate with respect to the $W_2$ distance between the training and test spaces. The error is the mean of 10 runs, and the error bars represent the one standard deviation.}
	\label{fig:extrapolation-W2}
\end{figure}

For both problems, we choose $l_{\text{train}}$ and $l_{\text{test}}$ in the range $\{0.1, 0.2, \dots, 1.0\}$. The training dataset size is chosen as 1000. The $L^2$ relative errors of DeepONets trained and tested on datasets with different $l_{\text{train}}$ and $l_{\text{test}}$ are shown in Figs.~\ref{fig:extrapolation-W2}A and B. When $l_{\text{train}} = l_{\text{test}}$ (Figs.~\ref{fig:extrapolation-W2}A and B, diagonal dash lines), the training and test functions are sampled from the same space, and thus it is interpolation and the error is smaller than $10^{-2}$. In the bottom right region where $l_{\text{train}} > l_{\text{test}}$, the error of DeepONet is larger. When $l_{\text{train}} < l_{\text{test}}$ (Figs.~\ref{fig:extrapolation-W2}A and B, left top region), although it is extrapolation, DeepONet still has a small error, i.e., DeepONet can predict accurately for smoother functions. Therefore, we have three scenarios:
\begin{equation*}
    \text{Prediction} =
    \begin{cases}
    \text{Interpolation (In.)}, & \text{when } l_{\text{train}} = l_{\text{test}}, \\
    \text{Ex.\textsuperscript{$-$}}, & \text{when } l_{\text{train}} < l_{\text{test}}, \\
    \text{Ex.\textsuperscript{+}}, & \text{when } l_{\text{train}} > l_{\text{test}}.
    \end{cases}
\end{equation*}
Here, the extrapolation for functions with smaller $l$ (i.e., less smoother) is denoted by Ex.\textsuperscript{+}, while extrapolation for functions with larger $l$ (i.e., more smoother) is denoted by Ex.\textsuperscript{$-$}.

\subsection{Extrapolation complexity and 2-Wasserstein distance}
We have tested DeepONet with different levels of extrapolation. To quantify the extrapolation complexity, we measure the distance between two GRFs using the 2-Wasserstein ($W_2$) metric~\cite{gelbrich1990formula}, as was suggested in Ref.~\cite{deeponetNatureML}. Let us consider two Gaussian processes $f_1 \sim \mathcal{GP}(m_1, k_1)$ and $f_2 \sim \mathcal{GP}(m_2, k_2)$ defined on a space $X$, where $m_1, m_2$ are the mean functions and $k_1, k_2$ are the covariance functions. A covariance functions $k_i$ is associated with a covariance operator $K_i:L^2(X)\rightarrow L^2(X)$ given by
$$
[K_i\phi](x)=\int_X k_i(x,s)\phi(s)ds, \quad \forall\phi \in L^2(X).
$$
Then the $W_2$ metric between the two GPs $f_1$ and $f_2$ is obtained as
$$
W_2(f_1, f_2) \coloneqq \left\{ \|m_1-m_2\|_2^2 + \text{Tr} \left[ K_1+K_2-2 \left(K_1^{\frac{1}{2}}K_2K_1^{\frac{1}{2}}\right)^{\frac{1}{2}} \right] \right\}^{\frac{1}{2}},
$$
where Tr is the trace. Following the definition, a larger $W_2$ distance represents a greater difference between two GRF spaces.

We show the testing error between each pair of GRFs with different correlation lengths in Figs.~\ref{fig:extrapolation-W2}A, B and C, and the relationship between the error and $W_2$ distance is shown in Figs.~\ref{fig:extrapolation-W2}D, E and F. We find that the extrapolation error grows with respect to the $W_2$ distance between the two GRF spaces. Specially, for the ODE problem, we have
\begin{equation} \label{eq:ode_convergence}
    \text{Error} \propto W_2^{1.7}.
\end{equation}
For the diffusion-reaction equation with the coefficient $k=0.01$, we have
\begin{equation} \label{eq:dr_convergence_0.01}
    \text{Error} \propto W_2^{2.1},
\end{equation}
and for the coefficient $k=0.5$,
\begin{equation} \label{eq:dr_convergence_0.5}
    \text{Error} \propto W_2^{1.9}.
\end{equation}
Therefore, the $W_2$ distance between the training and test spaces can be used as a measure of the extrapolation complexity.

In Eqs.~\eqref{eq:ode_convergence}, \eqref{eq:dr_convergence_0.01}, and \eqref{eq:dr_convergence_0.5}, the test errors of different problems converge with different rates. For the diffusion-reaction equation, the value of $k$ has an influence on the convergence rate. To further confirm if the influence of $k$ on the convergence rate is significant, we compute the 95\% confidence intervals for the two convergence rates above as $[2.01, 2.25]$ and $[1.79, 1.92]$, and the $p$-values of the two-sided T-test is $0.0001$, which implies the difference of the convergence rates is significant. Moreover, we observe that for a larger value of $k$, the extrapolation error grows slower, which is a surprising result since larger $k$ leads to a stronger nonlinearity in the PDE. This is a preliminary result, and further investigation is required in future work. 

\subsection{Understanding the extrapolation error}

To further understand the extrapolation error, we use the diffusion-reaction equation as an example and investigate several factors that contribute to the extrapolation error. Unless otherwise stated, we use the following hyperparameters. The DeepONet has one hidden layer for the branch net and two hidden layers for the trunk net, each with 100 neurons per layer. The activation function for both branch and trunk nets is ReLU. The correlation length of the training dataset is $l_{\text{train}} = 0.5$. In this case, $l_{\text{test}}<0.5$, $l_{\text{test}}>0.5$ and $l_{\text{test}}=0.5$ represents Ex.\textsuperscript{+}, Ex.\textsuperscript{$-$}, and interpolation, respectively. DeepONets are trained with the Adam optimizer with a learning rate of 0.001 for 500,000 iterations.

Our main findings of this subsection are as follows.
\begin{itemize}
    \item Section~\ref{sec:U-curve}: Similar to the classical U-shaped bias-variance trade-off curve for the test error of interpolation, Ex.\textsuperscript{+} also has a U-shaped curve with respect to the network size and the number of training iterations. Compared with interpolation, Ex.\textsuperscript{+} has larger error and earlier transition point.
    \item Section~\ref{sec:dataset_size}: Increasing training dataset size results in better accuracy for Ex.\textsuperscript{+}.
    \item Section~\ref{sec:activation}: Different activation functions perform differently for Ex.\textsuperscript{+}. The layer-wise locally adaptive activation functions (L-LAAF)~\cite{jagtap2020locally, jagtap2022deep} slightly outperform their corresponding non-adaptive activation functions.
\end{itemize}

\subsubsection{Bias-variance trade-off for extrapolation}
\label{sec:U-curve}

\begin{figure}[htbp]
    \centering
    \includegraphics[width=\textwidth]{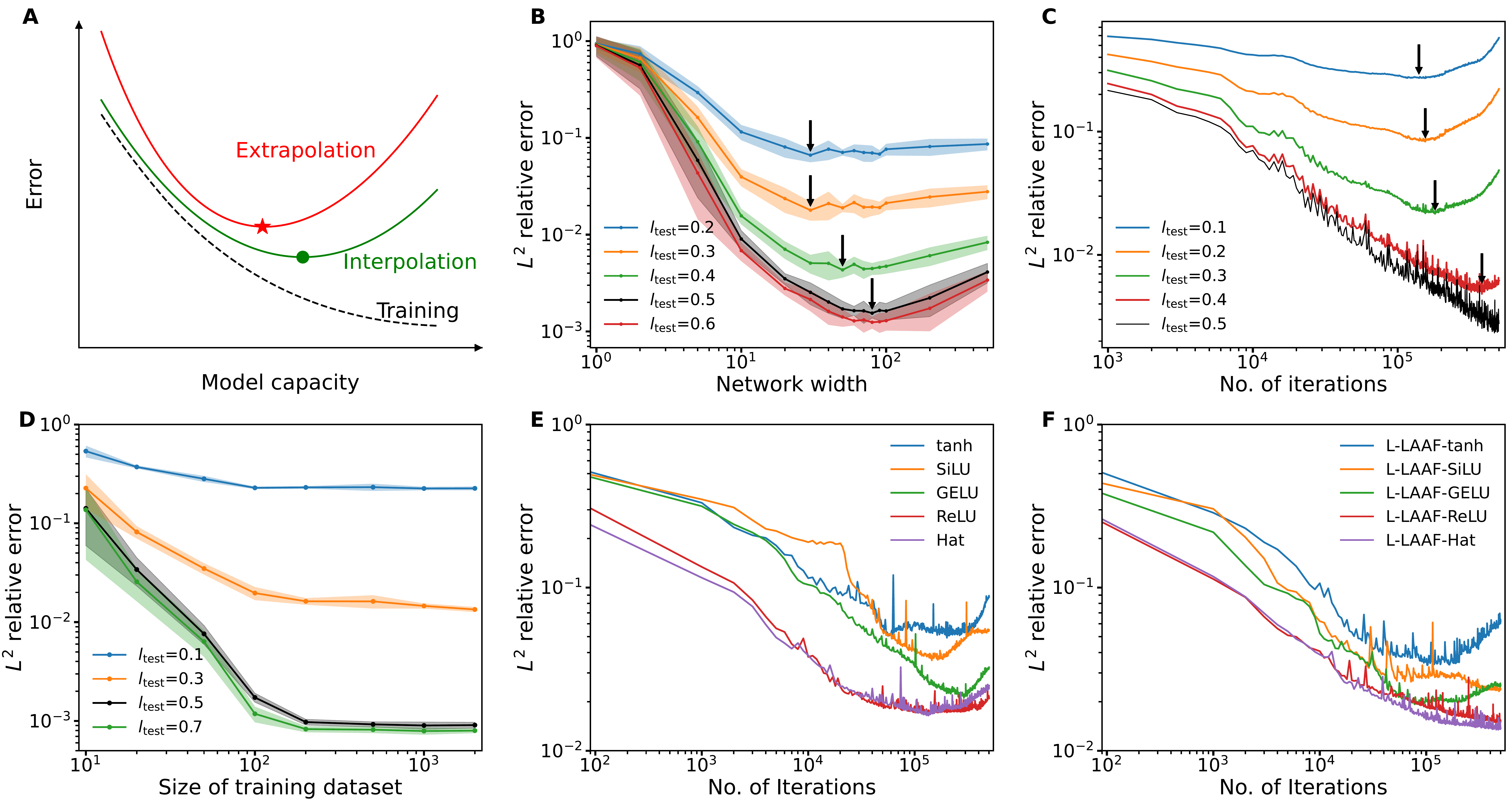}
    \caption{\textbf{Test error of Ex.\textsuperscript{+} for the diffusion-reaction equation.} (\textbf{A}) Schematic of typical curves for training error, interpolation test error, and Ex.\textsuperscript{+} test error versus model capacity. Test error has a U-shaped error curve. Red star and green dot represent the location where overfitting starts in Ex.\textsuperscript{+} and interpolation scenario, respectively. (\textbf{B}) The test error for different test datasets when using different network widths. The arrows indicate the transition point between underfitting and overfitting. The curves and shaded regions represent the mean and one standard deviation of 10 runs. (\textbf{C}) The test error for different test datasets during the training of DeepONet. The network width is 100. (\textbf{D}) The test error for different test datasets when using different training dataset sizes. (\textbf{E}) The test error when using different activation functions.  (\textbf{F}) The test error when using layer-wise locally adaptive activation functions (L-LAAF). For all the experiments, we use $l_{\text{train}} = 0.5$.}
    \label{fig:tendency}
\end{figure}

One of the central tenets in machine learning is the bias-variance trade-off~\cite{geman1992neural,hastie2009elements}, which implies a U-shaped curve for test error of interpolation as the model capacity grows, while the training error decreases monotonically (Fig.~\ref{fig:tendency}A). The bottom of the U-shaped curve is achieved at the transition point which balances the bias and variance. To the left of the transition point, the model is underfitting, and to the right, it is overfitting. There are several factors that affect the model capacity, such as network sizes and training iterations. Here, we show that the test error of Ex.\textsuperscript{+} also has a U-shaped curve.

To investigate the influence of the network size, we choose different network widths ranging from 1 to 500. We use a smaller training dataset of 100 so that we can observe the overfitting more easily. The correlation length of the test datasets is chosen from 0.2 to 0.6. For a fixed test dataset, we observe a U-shaped curve (Fig.~\ref{fig:tendency}B). Moreover, the transition point between underfitting and overfitting (indicated by the arrows in Fig.~\ref{fig:tendency}B) occurs earlier for Ex.\textsuperscript{+} than interpolation.

Next, we study the two phases of underfitting and overfitting during the training of DeepONets. We use five test datasets generated with different correlation lengths $l_{\text{test}}$ from 0.1 to 0.5. As shown in Fig.~\ref{fig:tendency}C, a smaller $l_{\text{test}}$ leads to a larger extrapolation error during the entire training process. Moreover, the transition point between underfitting and overfitting (indicated by the arrows in Fig.~\ref{fig:tendency}C) occurs earlier for smaller $l_{\text{test}}$.

Here, we have observed the classical U-shaped curves for the test errors of interpolation and extrapolation. However, recently a ``double-descent'' test error curve has been observed for neural networks~\cite{belkin2019reconciling,nakkiran2021deep}, i.e., when the model capacity is further increased, after a certain point the test error would decrease again. We do not observe the double-descent behavior for DeepONets, and one possible reason is that our model capacity is not large enough.

\subsubsection{Training dataset size}
\label{sec:dataset_size}

Next, we explore the effect of training dataset size in both interpolation and extrapolation cases. The dataset size ranges from 10 to 2000, and we show the test errors of four datasets in Fig.~\ref{fig:tendency}D. It is well known that a larger training dataset leads to better accuracy for interpolation. As shown in Fig.~\ref{fig:tendency}D, for extrapolation, a larger training dataset still leads to a smaller test error. However, the improvement exhibits a diminishing trend as the dataset size goes large.

\subsubsection{Activation function}
\label{sec:activation}

Another factor that the accuracy of DeepONet is contingent upon is the activation function. To evaluate the effect of activation functions, for the trunk nets, we consider four activation functions widely used in deep learning, including the hyperbolic tangent ($\tanh$), rectified linear units (ReLU, $\max(0,x)$)~\cite{nair2010rectified}, sigmoid linear units (SiLU, $\frac{x}{1+e^{-x}}$)~\cite{elfwing2018sigmoid}, and Gaussian error linear units (GELU, $x \cdot \frac{1}{2}\left[1+\text{erf} (x/\sqrt{2})\right]$)~\cite{hendrycks2016gaussian}. In addition, we explore the Hat activation function~\cite{hong2022activation} defined by
\begin{equation*}
    \text{Hat}(x) \coloneqq
    \begin{cases}
    0, & x < 0 \text{ or } x \geq 2, \\
    x, & 0 \leq x < 1, \\
    2-x, & 1 \leq x < 2.
    \end{cases}
\end{equation*}
For branch nets, we still use ReLU. The correlation lengths for training and testing datasets are $l_{\text{train}} = 0.5$ and $l_{\text{test}}=0.3$, respectively. ReLU and Hat have similar results and outperform the rest of the activation functions (Fig.~\ref{fig:tendency}E). GELU also presents a satisfactory performance despite the overfitting after a certain number of training iterations.

Besides these commonly used activation functions, we also consider the layer-wise locally adaptive activation functions (L-LAAF)~\cite{jagtap2020locally, jagtap2022deep} in the form of $\sigma(n \cdot a \cdot x)$, where $\sigma$ is a standard activation function, $n$ is a scaling factor, and $a$ is a trainable parameter. L-LAAF introduces the trainable parameter in each layer, thus leading to a local adaptation of activation function. The performance of L-LAAF depends on the scaling factor $n$, so we performed a grid search for $n$ from 1, 2, 5, and 10 (Appendix Fig.~\ref{fig:L-LAAF}). With other settings remaining the same, L-LAAF with the optimal $n$ slightly outperforms non-adaptive activation functions (Fig.~\ref{fig:tendency}F). L-LAAF-Hat and L-LAAF-ReLU perform the best among the five activation functions, and most importantly the tendency of overfitting is alleviated.

\section{Reliable learning methods for safe extrapolation}
\label{sec:methods}

As we demonstrate in Section~\ref{sec:extp}, compared with interpolation, extrapolation leads to a much larger prediction error. However, extrapolation is usually unavoidable in real applications. In this section, we propose several reliable learning methods to guarantee a safe prediction in the extrapolation region.

\subsection{Workflow of extrapolation}

We continue to consider the setup of operator learning in Section~\ref{sec:deeponet}. We assume that we have a training dataset $\mathcal{T}$, and then we train a DeepONet with $\mathcal{T}$. We denote this pre-trained DeepONet by $\mathcal{G}_{\bm{\theta}}$, where $\bm{\theta}$ is the set of trainable parameters in the network. If the size of $\mathcal{T}$ is large enough and the DeepONet is well trained, then the pre-trained DeepONet can have accurate predictions for interpolation. In this study, we do not develop new methods to improve the interpolation performance. Our goal is to have an accurate prediction for a new input function $v$, irrespective if $v$ belongs to interpolation or extrapolation.

In general, it is very difficult to have an accurate prediction for extrapolation~\cite{158898,jin2020quantifying,xu2020neural}, e.g., see the simple problem of function regression in Fig.~\ref{fig:fig_1}A and the examples of DeepONets in Refs.~\cite{deeponetNatureML,lin2021operator,MAO2021110698,kissas2022learning,yin2022interfacing}. Hence, in order to reduce the extrapolation error, it is essential to have additional information. Here, we consider the following two scenarios and develop corresponding methods to address the extrapolation problem.
\begin{enumerate}
    \item Physics (Section~\ref{sec:physics}): We know the governing PDEs and/or the physical constraints of the system:
    \begin{equation*}
        \mathcal{F}[u; v] = 0
    \end{equation*}
    with suitable initial and boundary conditions
    $$\mathcal{B}[u; v] = 0.$$
    \item Sparse observations (Sections~\ref{sec:fine-tune} and \ref{sec:mf}): We have sparse observations of the output function $u = \mathcal{G}(v)$ at $N_{\text{obs}}$ locations:
    \begin{equation} \label{eq:data}
        \mathcal{D} = \left\{ (\xi_1, u(\xi_1)), (\xi_2, u(\xi_2)), \dots, (\xi_{N_{\text{obs}}},u(\xi_{N_{\text{obs}}})) \right\}.
    \end{equation}
\end{enumerate}

\begin{figure}[htbp]
    \centering
    \includegraphics[width=\textwidth]{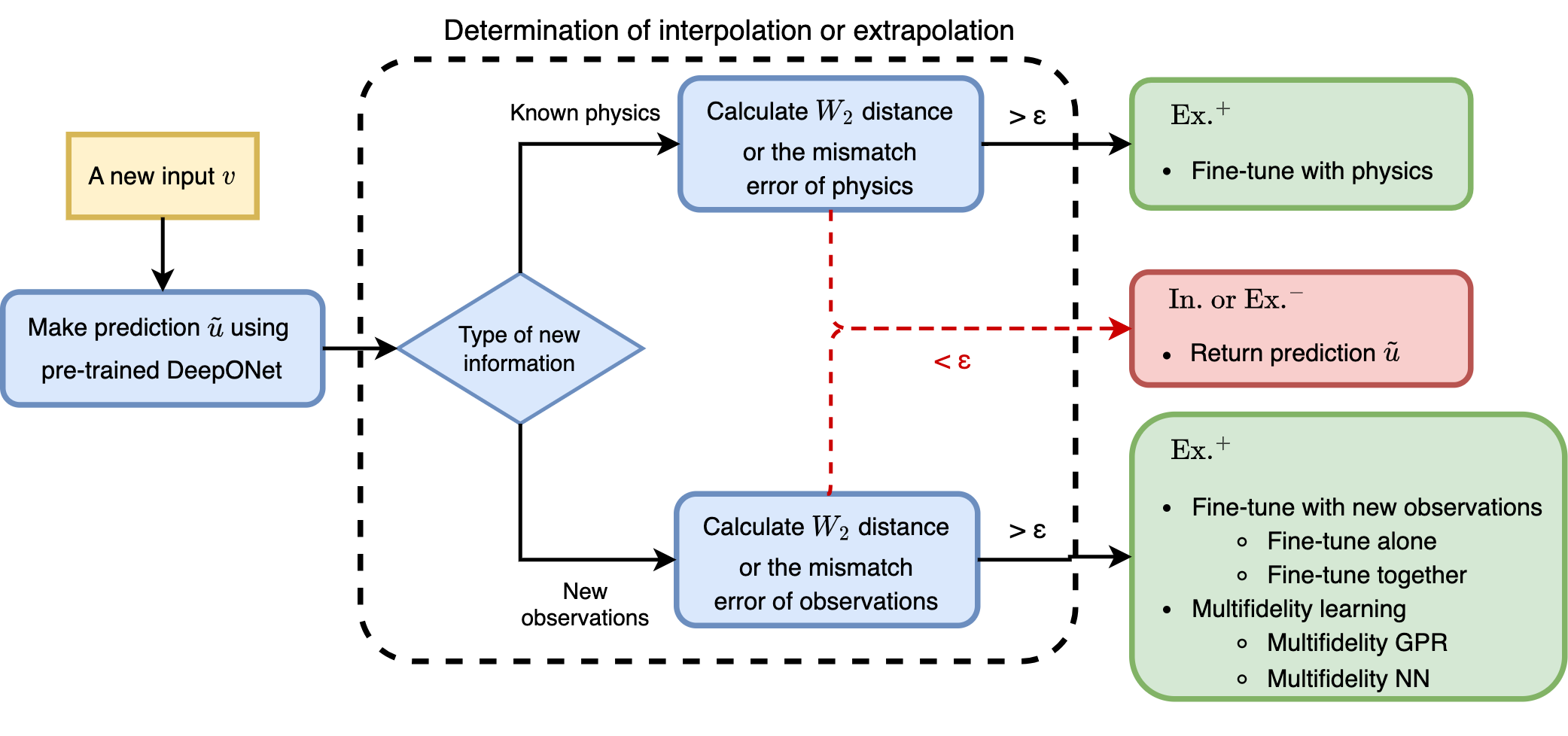}
    \caption{\textbf{Flowchart of predicting the output $\tilde{u}$ for an input $v$.} For extrapolation, we use one of the following learning methods: fine-tune with physics, fine-tune with sparse new observations, or multifidelity learning. Here, $\epsilon$ denotes a user specified threshold to determine interpolation or extrapolation.}
    \label{fig:flow-chart}
\end{figure}

Since we aim to handle any input function $v$, the first step is to determine if $v$ is in the region of interpolation or extrapolation, which is discussed in Section~\ref{sec:determination}. If it is interpolation, then it becomes trivial, i.e., we only need to predict the output by the trained DeepONet; otherwise, we need to extrapolate via the proposed methods. The entire workflow is shown in Fig.~\ref{fig:flow-chart}.

\subsection{Determination of interpolation or extrapolation}
\label{sec:determination}

One approach to determine interpolation or extrapolation is to compare the new function $v$ with the functions in the training dataset. If $v$ is less smooth than the training functions, then it is extrapolation, otherwise it is interpolation as we presented in Section~\ref{sec:extp}. Here, because we have extra information of either physics or observations, we first predict the output $\tilde{u}$ by the pre-trained DeepONet $\mathcal{G}_{\bm{\theta}}$, i.e., $\tilde{u} = \mathcal{G}_{\bm{\theta}}(v)$, and then check if the $\tilde{u}$ is consistent with the physics $\mathcal{F}$ or observations $\mathcal{D}$.

Specifically, we propose to compute one of the following errors of mismatch:
\begin{itemize}
    \item Mismatch error of physics, i.e., the mean PDE residual:
    \begin{equation*}
        \mathcal{E}_{\text{phys}} = \frac{1}{\text{Area}(\Omega)} \int_\Omega |\mathcal{F}[\tilde{u}; v]| d\xi,
    \end{equation*}
    where $\Omega$ is the domain of the PDE.
    \item Mismatch error of observations, i.e., the root relative squared error (RRSE):
    \begin{equation*}
        \mathcal{E}_{\text{obs}} = \sqrt{\frac{\sum_{i=1}^{N_{\text{obs}}} (\tilde{u}(\xi_i) - u(\xi_i))^2}{\sum_{i=1}^{N_{\text{obs}}} u^2(\xi_i)}}.
    \end{equation*}
\end{itemize}
To compute $\mathcal{F}[\tilde{u}; v]$ in $\mathcal{E}_{\text{phys}}$, we need the derivatives of the DeepONet output $\tilde{u}$ with respect to the trunk net input $\xi$ while the branch net input is fixed at $v$. This can be computed via automatic differentiation as was done in physics-informed neural networks (PINNs) and physics-informed DeepONets~\cite{wang2021learning,GOSWAMI2022114587}.

\begin{figure}[htbp]
    \centering
    \includegraphics[width=\textwidth]{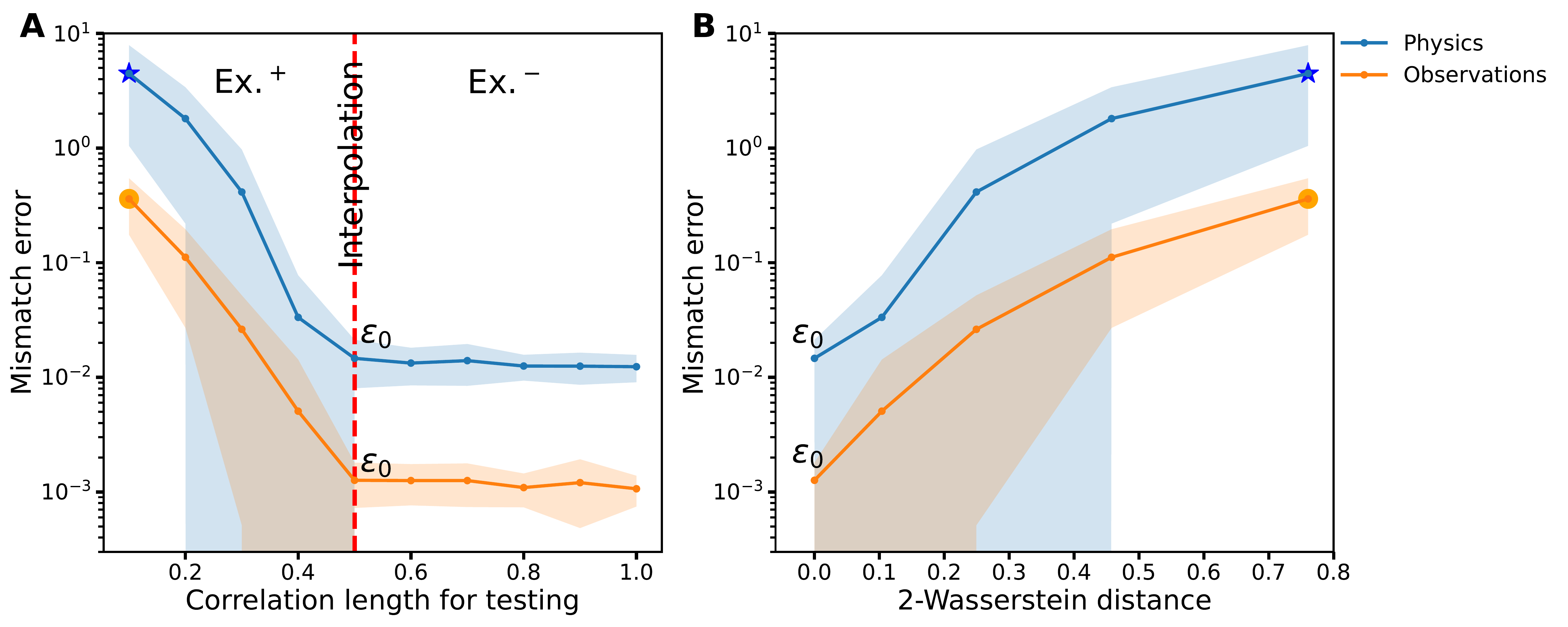}
    \caption{\textbf{Mismatch error of physics ($\mathcal{E}_{\text{phys}}$) or observations ($\mathcal{E}_{\text{obs}}$) for the diffusion-reaction equation}. (\textbf{A}) Mismatch error for different testing correlation length. (\textbf{B}) Mismatch error for different 2-Wasserstein distance. The correlation length for training is 0.5. The number of new observations for $\mathcal{E}_{\text{obs}}$ is 100. The curves and shaded regions represent the mean and one standard deviation of 100 test functions.}
    \label{fig:threshold}
\end{figure}

To verify that $\mathcal{E}_{\text{phys}}$ and $\mathcal{E}_{\text{obs}}$ are good metrics of interpolation and extrapolation, we take the diffusion-reaction equation as an example. We train a DeepONet with the functions sampled from GRF with $l_{\text{train}} = 0.5$, and then compute $\mathcal{E}_{\text{phys}}$ and $\mathcal{E}_{\text{obs}}$ for different functions randomly sampled from GRFs with different correlation lengths $l$. We denote the value of $\mathcal{E}_{\text{phys}}$ or $\mathcal{E}_{\text{obs}}$ for $l=l_{\text{train}}$ by $\epsilon_0$. In the interpolation and Ex.\textsuperscript{$-$} regions, i.e., $l \ge l_{\text{train}}$, the value of $\mathcal{E}_{\text{phys}}$ or $\mathcal{E}_{\text{obs}}$ is close to $\epsilon_0$ (Fig.~\ref{fig:threshold}A). In the Ex.\textsuperscript{+} region, i.e., $l < l_{\text{train}}$, the value of $\mathcal{E}_{\text{phys}}$ or $\mathcal{E}_{\text{obs}}$ is much larger than $\epsilon_0$. Moreover, for two spaces with larger 2-Wasserstein distance, the mismatch error is also larger (Fig.~\ref{fig:threshold}B). Therefore, we can select the threshold $\epsilon = \alpha \epsilon_0$, where $\alpha \ge 1$ is a user specified tolerance factor, and then compare $\mathcal{E}_{\text{phys}}$ or $\mathcal{E}_{\text{obs}}$ with $\epsilon$. If $\mathcal{E}_{\text{phys}} > \epsilon$ or $\mathcal{E}_{\text{obs}} > \epsilon$, then it is Ex.\textsuperscript{+}; otherwise, it is interpolation or Ex.\textsuperscript{$-$} (Fig.~\ref{fig:flow-chart}).

\subsection{Extrapolation via fine-tuning with physics (FT-Phys)}
\label{sec:physics}

We first discuss how to extrapolate with the additional information of physics (Algorithm~\ref{alg:physics}). For a new input $v$ in the extrapolation region, the prediction $\tilde{u}$ by the pre-trained DeepONet $\mathcal{G}_{\bm{\theta}}$ may not satisfy the PDE. We propose to fine-tune $\mathcal{G}_{\bm{\theta}}$ to minimize the loss
\begin{equation} \label{eq:physics}
    \mathcal{L}_{\text{phys}} = w_{\mathcal{F}} \mathcal{L}_{\mathcal{F}} + w_{\mathcal{B}} \mathcal{L}_{\mathcal{B}},
\end{equation}
where $\mathcal{L}_{\mathcal{F}}$ is the loss of PDE residuals
$$\mathcal{L}_{\mathcal{F}} = \frac{1}{|\mathcal{T}_{\mathcal{F}}|} \sum_{\xi \in \mathcal{T}_{\mathcal{F}}} |\mathcal{F}[\mathcal{G}_{\bm{\theta}}(v)(\xi); v]|^2,$$
$\mathcal{L}_{\mathcal{B}}$ is the loss of initial and boundary conditions
$$\mathcal{L}_{\mathcal{B}} = \frac{1}{|\mathcal{T}_{\mathcal{B}}|} \sum_{\xi \in \mathcal{T}_{\mathcal{B}}} |\mathcal{B}[\mathcal{G}_{\bm{\theta}}(v)(\xi); v]|^2,$$
and $w_{\mathcal{F}}$ and $w_{\mathcal{B}}$ are the weights. $\mathcal{T}_{\mathcal{F}}$ and $\mathcal{T}_{\mathcal{B}}$ are two sets of points sampled in the domain and on the initial and boundary locations, respectively. A DeepONet has two subnetworks (a branch net and a trunk net), and thus we could choose to fine-tune the entire DeepONet or only one subnetwork, while other parts remain unchanged. Moreover, as each subnetwork has multiple layers, we can also only fine-tune certain layers of the subnetwork. In this study, we consider four approaches: (1) fine-tuning both trunk and branch nets (Branch \& Trunk), (2) fine-tuning the branch net, (3) fine-tuning the trunk net, and (4) fine-tuning the last layer of the trunk net (Trunk last).

\begin{algorithm}[htbp]
\caption{\textbf{Extrapolation via fine-tuning with physics.}}
\label{alg:physics}
\KwIn{A pre-trained DeepONet $\mathcal{G}_{\bm{\theta}}$, and a new input function $v$}
\KwData{Physics information $\mathcal{F}$ and $\mathcal{B}$}
\KwOut{Prediction $u = \mathcal{G}_{\bm{\theta}}(v)$}
Fix the input of the branch net of $\mathcal{G}_{\bm{\theta}}$ to be $v$\;
Sample the two sets of training points $\mathcal{T}_{\mathcal{F}}$ and $\mathcal{T}_{\mathcal{B}}$ for the equation and initial/boundary conditions\;
Select the weights $w_{\mathcal{F}}$ and $w_{\mathcal{B}}$\;
Train all the parameters or a subset of the parameters of $\mathcal{G}_{\bm{\theta}}$ by minimizing the loss $\mathcal{L}_{\text{phys}}$ of Eq.~\eqref{eq:physics}\;
\end{algorithm}

The idea of first pre-training and then fine-tuning is also relevant to the field of transfer learning (TL)~\cite{pan2009survey,zhuang2020comprehensive,deng2021deep}, which aims to extract the knowledge from one or more source tasks and then apply the knowledge to a target task. However, most TL methods are developed to deal with covariate shift, label shift (or prior shift, target shift), and conditional shift, but not extrapolation.

As the branch net input is fixed at $v$, this fine-tuning method works in the same way as PINNs~\cite{RAISSI2019686,lu2021deepxde}. Compared with a PINN with random initialization, a pre-trained DeepONet provides a better initialization, which makes the training much faster, especially when the extrapolation complexity is small. The proposed method also uses the same technique as physics-informed DeepONets~\cite{wang2021learning,GOSWAMI2022114587}, but physics-informed DeepONets still require the target function space for training.

\subsection{Extrapolation via fine-tuning with sparse new observations}
\label{sec:fine-tune}

With the information of new sparse observations $\mathcal{D}$ in Eq.~\eqref{eq:data}, we propose a few different approaches for Ex.\textsuperscript{+} scenario. Similar to the fine-tuning with physics, the first proposed method is also in the spirit of transfer learning, i.e., we fine-tune the pre-trained DeepONet with the new observations (Algorithm~\ref{alg:tf-data}).

\paragraph{Fine-tune with new observations alone (FT-Obs-A).}
As the pre-trained DeepONet is not consistent with the ground truth observations, we fine-tune the DeepONet to better fit the observations. Specifically, we further train the DeepONet by minimizing the mean squared error (MSE)
\begin{equation} \label{eq:ft-obs-a}
    \mathcal{L}_{\text{obs}} = \frac{1}{N_{\text{obs}}}\sum_{i=1}^{N_{\text{obs}}} (\mathcal{G}_{\bm{\theta}}(v)(\xi_i) - u(\xi_i))^2.
\end{equation}
In this approach, we only use the new observations for fine-tuning, and thus we call it ``fine-tune alone'' to distinguish from the next approach.

\begin{algorithm}[htbp]
\caption{\textbf{Extrapolation via fine-tuning with new observations.}}
\label{alg:tf-data}
\KwIn{A pre-trained DeepONet $\mathcal{G}_{\bm{\theta}}$, and a new input function $v$}
\KwData{New observations $\mathcal{D}$}
\KwOut{Prediction $u = \mathcal{G}_{\bm{\theta}}(v)$}
Fix the input of the branch net of $\mathcal{G}_{\bm{\theta}}$ to be $v$\;
Fine-tune $\mathcal{G}_{\bm{\theta}}$ by minimizing the loss of Eq.~\eqref{eq:ft-obs-a} for fine-tuning alone or Eq.~\eqref{eq:ft-obs-t} for fine-tuning together\;
\end{algorithm}

The approach above is simple and has low computational cost. However, as $N_{\text{obs}}$ is usually very small, it may have the issue of overfitting. It may also have the issue of catastrophic forgetting~\cite{goodfellow2013empirical,kirkpatrick2017overcoming}, i.e., DeepONet would forget previously learned information upon learning the new observations. Considering the fact that both the training dataset and the new observations satisfy the same operator $\mathcal{G}$, they can be learned by the same DeepONet at the same time. Hence, we propose the following fine-tuning approach to prevent overfitting and catastrophic forgetting.

\paragraph{Fine-tune with training data and new observations together (FT-Obs-T).}
Instead of fitting DeepONet with only the new observations, we fine-tune the pre-trained DeepONet with new observed data together with the original training dataset $\mathcal{T}$ via the loss
\begin{equation} \label{eq:ft-obs-t}
    \mathcal{L}_{\mathcal{T},\text{obs}} = \mathcal{L}_{\mathcal{T}} + \lambda \mathcal{L}_{\text{obs}}
\end{equation}
where $\mathcal{L}_{\mathcal{T}}$ is the original loss for $\mathcal{T}$, e.g., MSE loss, and $\lambda$ is a weight.

Compared with the FT-Obs-A above, FT-Obs-T keeps learning from $\mathcal{T}$ via the loss of $\mathcal{L}_{\mathcal{T}}$, which mitigates the problem of catastrophic forgetting. Moreover, $\mathcal{L}_{\mathcal{T}}$ has an effect of regularization to prevent the overfitting of new sparse observations. By tuning the value of $\lambda$, we can balance remembering the original information $\mathcal{T}$ and learning from the new information $\mathcal{D}$. 

\subsection{Extrapolation via multifidelity learning with sparse new observations}
\label{sec:mf}

In all the previous methods, we use the idea of fine-tuning. Here, we propose a different approach based on multifidelity learning~\cite{kennedy2000predicting}. The idea of multifidelity learning is that instead of learning from a large dataset of high accuracy (i.e., high fidelity), we only use a small high-fidelity dataset complemented by another dataset of low accuracy (i.e., low fidelity). Specifically, to predict $\mathcal{G}(v)$ for the new function $v$, the sparse new observations $\mathcal{D}$ is the high-fidelity dataset, while the prediction $\tilde{u} = \mathcal{G}_{\bm{\theta}}(v)$ from the pre-trained DeepONet is the low-fidelity dataset.

We integrate high- and low-fidelity datasets via two multifidelity methods: multifidelity Gaussian process regression (MFGPR)~\cite{sobester2008engineering} or multifidelity neural networks (MFNN)~\cite{meng2020composite,lu2020extraction,lu2022machine}. In MFGPR, we model the high- and low-fidelity functions by Gaussian processes with the radial-basis function kernel in Eq.~\eqref{eq:rbf}. Then, the model is trained by minimizing the negative log marginal likelihood on the datasets. In MFNN, we have one fully-connected neural network to learn the low-fidelity dataset and another fully-connected neural network to learn the correlation between low- and high-fidelity. For more details of MFNN, see Refs.~\cite{meng2020composite,lu2020extraction}. The algorithm is shown in Algorithm~\ref{alg:mf}.

\begin{algorithm}
\caption{\textbf{Extrapolation via multifidelity learning with new observations.}}
\label{alg:mf}
\KwIn{A pre-trained DeepONet $\mathcal{G}_{\bm{\theta}}$, and a new input function $v$}
\KwData{New observations $\mathcal{D}$}
\KwOut{Prediction $u$}
Compute the prediction $\tilde{u} = \mathcal{G}_{\bm{\theta}}(v)$\;
Sample a set of dense points $\mathcal{S} = \{\xi_i\}_{i=1}^{|\mathcal{S}|}$ and use $\{(\xi_i, \tilde{u}(\xi_i))\}_{i=1}^{|\mathcal{S}|}$ as the low-fidelity dataset\;
Use $\mathcal{D}$ as the high-fidelity dataset\;
Train a multifidelity model (MFGPR or MFNN) on the multifidelity datasets\;
\end{algorithm}

\section{Extrapolation results}
\label{sec:results}

In this section, we test the proposed extrapolation methods with different PDEs. The hyperparameters used in this study can be found in Appendix~\ref{sec:hyperparameter}. For all experiments, the Python library DeepXDE~\cite{lu2021deepxde} is utilized to implement the algorithms and train the neural networks. The code in this study is publicly available from the GitHub repository \url{https://github.com/lu-group/deeponet-extrapolation}.

To demonstrate the effectiveness of the extrapolation methods, we consider three different baselines. We use the pre-trained DeepONet as the first baseline to show the Ex.\textsuperscript{+} error without any additional information. Physics-informed DeepONet (PIDeepONet)~\cite{pideeponet} is taken as another baseline model, which is trained by the PDE loss and initial/boundary conditions. When we know the governing physics, we also use PINN as the baseline for the method of fine-tuning with physics. The network size and activation function of PINN are the same as those of trunk net of DeepONets. When we have extra new observations, we perform a Gaussian process regression (GPR) with the observations as a strong single-fidelity baseline for fine-tuning with observations and multifidelity learning. The new observations are randomly sampled in the domain unless otherwise stated.

\subsection{Antiderivative operator}
\label{subsec:ode}

First, we consider the antiderivative operator in Section~\ref{subsec:2.2}. The goal is to learn the operator mapping from $v(x)$ to the solution $u(x)$. For the training dataset, $v(x)$ is sampled from GRF of an RBF kernel with the correlation length $l_{\text{train}}=0.5$. To test the Ex.\textsuperscript{+}, we generate a test dataset of 100 functions with $l_{\text{test}}=0.2$.

\begin{figure}[htbp]
    \centering
    \includegraphics[width=\textwidth]{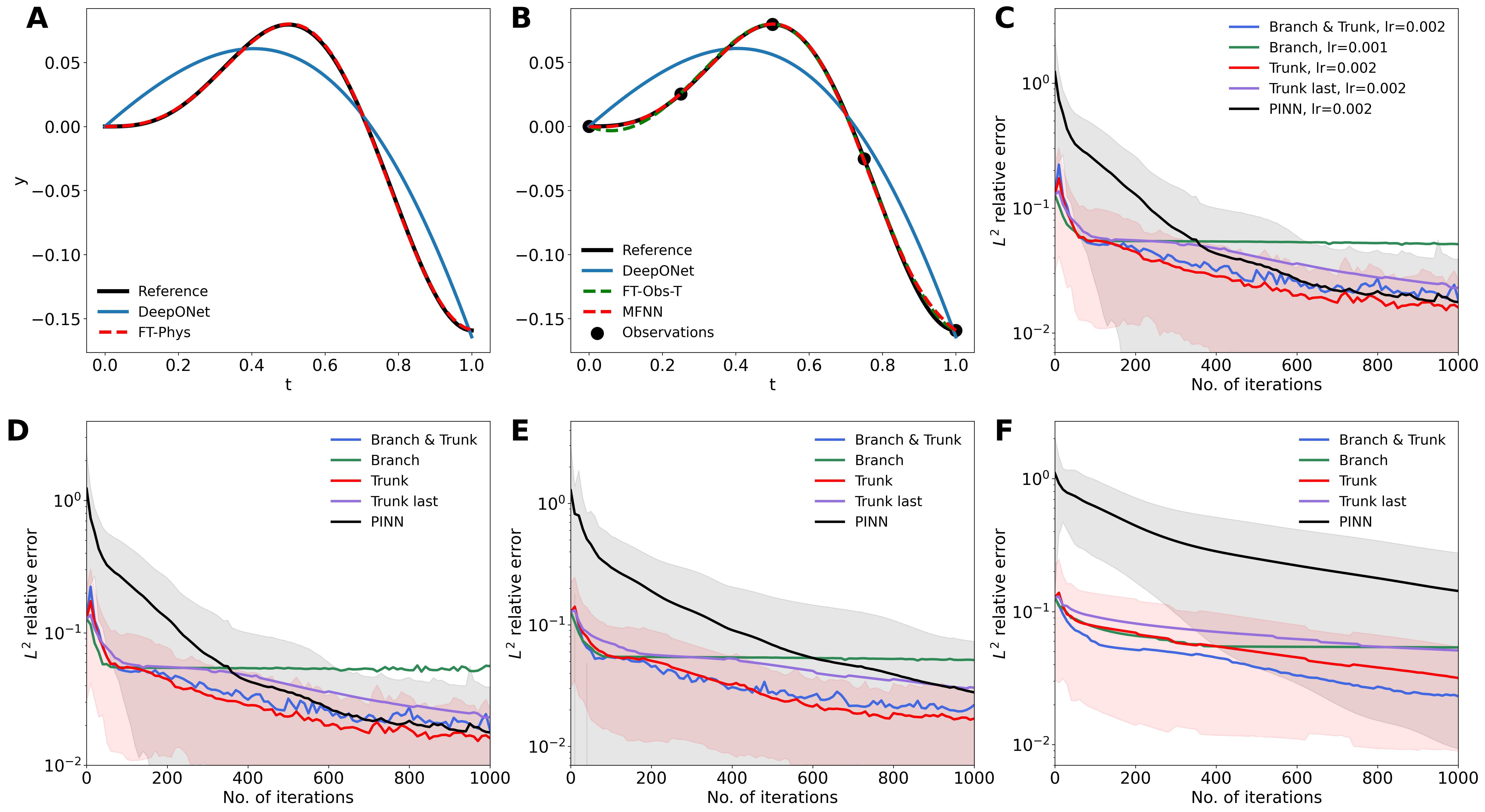}
     \caption{\textbf{Antiderivative operator in Section~\ref{subsec:ode}.} (\textbf{A}) Predictions of the pre-trained DeepONet and fine-tuning with physics. (\textbf{B}) Predictions of the pre-trained DeepONet, FT-Obs-T, and MFNN with 5 new observations. (\textbf{C}) The best result of each method among different learning rates. (\textbf{D}, \textbf{E}, and \textbf{F}) Training trajectories under the learning rate of (D) 0.002, (E) 0.001, and (F) 0.0002. The curves and shaded regions represent the mean and one standard deviation of 100 test cases. FT-Phys and MFNN obtain the best results. (For clarity, only some standard deviations are plotted.)}
    \label{fig:ode}
\end{figure}

Table~\ref{table:ode} summarizes the results of different methods. The pre-trained DeepONet has an average $L^2$ relative error of 11.6\%, and PIDeepONet has an average $L_2$ relative error of 7.44\% for the test dataset. When we have information of physics, fine-tuning with physics achieves accuracy of 1.52\%, which is more accurate than PINN. When we have more than 5 sparse observations, FT-Obs-T, MFGPR and MFNN achieve accuracy about 2\%, but FT-Obs-A has a relatively large error. In this case, multifidelity learning (MFGPR and MFNN) outperforms FT-Obs-A and FT-Obs-T. We note that this example is a relatively simple problem and mulifidelity methods work well, but for the other examples, fine-tuning with new observations has better accuracy. Moreover, an example of prediction result is also provided in Figs.~\ref{fig:ode}A and B to illustrate the enhancement resulted from employing proposed methods. While the performance of FT-Obs-A, FT-Obs-T, MFNN, and MFGP are similar, only FT-Obs-T and MFNN are plotted. Irrespective of new physical information (Fig.~\ref{fig:ode}A) or new observations (Fig.~\ref{fig:ode}B), the prediction results confirm that the proposed methods ameliorate the effect of Ex.\textsuperscript{+} to a great extent.

\begin{table}[htbp]
    \centering
    \caption{\textbf{$L^2$ relative error of different methods for the antiderivative operator in Section~\ref{subsec:ode}.} $l_{\text{train}}=0.5$ and $l_{\text{test}}=0.2$. Bold font indicates the smallest two errors in each case, and the underlined text indicates the smallest error.}
    \label{table:ode}
    \begin{tabular}{lcccc}
    \toprule
    DeepONet & \multicolumn{4}{c}{ Error(In.): $0.93\pm0.20\%$ \quad Error(Ex.\textsuperscript{+}): $11.6 \pm 5.26\%$} \\
    PIDeepONet & \multicolumn{4}{c}{ Error(In.): $0.81\pm0.69\%$ \quad Error(Ex.\textsuperscript{+}): $17.4 \pm 6.01\%$} \\
    \midrule
    \midrule
    PINN & \multicolumn{4}{c}{1.77 $\pm$ 2.14\%} \\
    \hdashline
      & Branch \& Trunk & Branch & Trunk & Trunk last \\
     FT-Phys & $\pmb{1.84}\pm\pmb{1.49}$\% & $5.09\pm3.66$\% & \underline{$\pmb{1.52}\pm\pmb{1.00}$}\% & $2.29\pm1.65$\% \\
     \midrule
     \midrule
     & 4 points & 5 points & 6 points & 7 points\\
    \hdashline
    GPR & $14.7 \pm 16.7 \%$ & $8.63 \pm 14.1 \%$ & $4.98\pm 12.8\%$  &  $3.15\pm 12.4\%$ \\
    \hdashline
    FT-Obs-A & $10.7\pm7.70\%$ & $5.98\pm4.15 \% $& $4.45\pm3.30\%$ & $4.22\pm3.11 \%$\\
    FT-Obs-T & $\pmb{9.05} \pm \pmb{6.50}$\% & $5.61\pm4.19 \%$ & $2.92\pm2.17 \%$ & $1.87\pm1.24\% $\\
    MFGPR & $13.0\pm11.2 \%$ & $\pmb{5.17} \pm \pmb{3.74}$\% & $\pmb{2.49} \pm \pmb{1.71}$\% & \underline{$\pmb{1.15} \pm \pmb{0.99}$}\% \\
    MFNN & \underline{$\pmb{8.99} \pm \pmb{5.83}$}\% & \underline{$\pmb{4.79} \pm \pmb{3.02}$}\% & \underline{$\pmb{2.10} \pm \pmb{1.33}$}\% & $\pmb{1.35} \pm \pmb{0.99}$\%\\
    \bottomrule
    \end{tabular}
\end{table}

\paragraph{Detailed results of FT-Phys.} For fine-tuning with physics, we test three different learning rates (i.e., 0.002, 0.001, 0.0002). For both PINN and fine-tuning with physics, we enforce the hard constraint for the initial condition. Among all variants of FT-Phys, fine-tuning the trunk net with a learning rate of 0.002 performs the best (Fig.~\ref{fig:ode}C). The performance of fine-tuning with physics for different learning rates is exhibited in Figs.~\ref{fig:ode}D, E, and F. Fine-tuning the trunk net and the entire DeepONet perform similarly well when the learning rate is large, and the full net slightly outperforms the trunk net as the learning rate is as small as 0.0002.

\subsection{Diffusion-reaction equation}
\label{subsec:diffusion-reaction}

Next, we consider the diffusion-reaction equation in Section~\ref{subsec:2.2}. We aim to train a DeepONet to learn the operator mapping from the source term $v(x)$ to the solution $u(x, t)$. The source term $v(x)$ is randomly sampled from a GRF with an RBF kernel with the correlation length $l_{\text{train}}=0.5$. To test the Ex.\textsuperscript{+}, we generate a test dataset of 100 functions with $l_{\text{test}}=0.2$.

Table~\ref{table:diffusion-reaction} summarizes the $L^2$ relative errors of different methods. The FT-Phys has the lowest errors, and the FT-Obs-T works the best among methods that require new observations. The pre-trained DeepONet has an average $L^2$ relative error of 10.4\%, and PIDeepONet has an average $L_2$ relative error of 10.2\% for the test dataset. As shown in Table~\ref{table:diffusion-reaction}, when we have physics, fine-tuning with trunk net at a learning rate of 0.002 can achieve low $L^2$ relative errors (0.32\%) and significantly outperform the pre-trained DeepONet and GPR. When we have 20, 50, and 100 new observations, FT-Obs-A and FT-Obs-T generally outperform MFGPR and MFNN. However, when we have 200 new observations, GPR and MFGPR reach low $L^2$ relative errors because the diffusion-reaction equation is relatively simple with a smooth solution, and 200 points are sufficient for MFGPR and GPR to obtain accurate results.

\begin{table}[htbp]
    \centering
    \caption{\textbf{$L^2$ relative error of different methods for the diffusion-reaction equation in Section~\ref{subsec:diffusion-reaction}.} $l_{\text{train}}=0.5$ and $l_{\text{test}}=0.2$. Bold font indicates the smallest two errors in each case, and the underlined text indicates the smallest error.}
    \label{table:diffusion-reaction}
    \begin{tabular}{lcccc}
    \toprule
    DeepONet & \multicolumn{4}{c}{ Error(In.): $0.74\pm0.29\%$ \quad Error(Ex.\textsuperscript{+}): $10.4\pm6.24\%$} \\
    PIDeepONet & \multicolumn{4}{c}{ Error(In.): $0.42\pm0.20\%$ \quad Error(Ex.\textsuperscript{+}):  $10.2\pm6.30\%$} \\
    \midrule
    \midrule
    PINN & \multicolumn{4}{c}{$0.49 \pm 0.24$\%} \\
    \hdashline
    & Branch \& Trunk & Branch & Trunk & Trunk last \\
    FT-Phys & $\pmb{0.37}\pm\pmb{0.16}$\% & $1.28\pm0.62$\% & \underline{$\pmb{0.32}\pm\pmb{0.15}$}\% & $0.48\pm0.21$\% \\
    \midrule
    \midrule
    & 20 points & 50 points & 100 points & 200 points\\
    \hdashline
    GPR & $34.5 \pm 15.0 \%$ & $9.63 \pm 4.55 \%$ & $2.59\pm 1.57\%$  &  $\pmb{0.61} \pm \pmb{0.39}$\% \\
    \hdashline
    FT-Obs-A & $\pmb{5.51} \pm \pmb{3.08}\%$ & $\pmb{3.36} \pm \pmb{1.72} \% $& $2.41\pm1.19\%$ & $1.63\pm0.69 \%$\\
    FT-Obs-T & \underline{$\pmb{4.56} \pm \pmb{2.66}$}\% & \underline{$\pmb{2.69} \pm \pmb{1.51}$}\% & \underline{$\pmb{1.83} \pm \pmb{0.86}$}\% & $1.32 \pm 0.59\% $\\
    MFGPR & $12.2\pm5.07 \%$ & $7.96\pm3.45 \%$ & $\pmb{2.26} \pm \pmb{1.49} \%$ & \underline{$\pmb{0.48} \pm \pmb{0.27}$}\% \\
    MFNN & $7.86\pm5.18 \%$ & $4.50\pm2.20 \%$ & $2.73\pm1.18\%$ & $1.50\pm0.68\%$\\
    \bottomrule
    \end{tabular}
\end{table}

\begin{figure}[htbp]
    \centering
    \includegraphics[width=\textwidth]{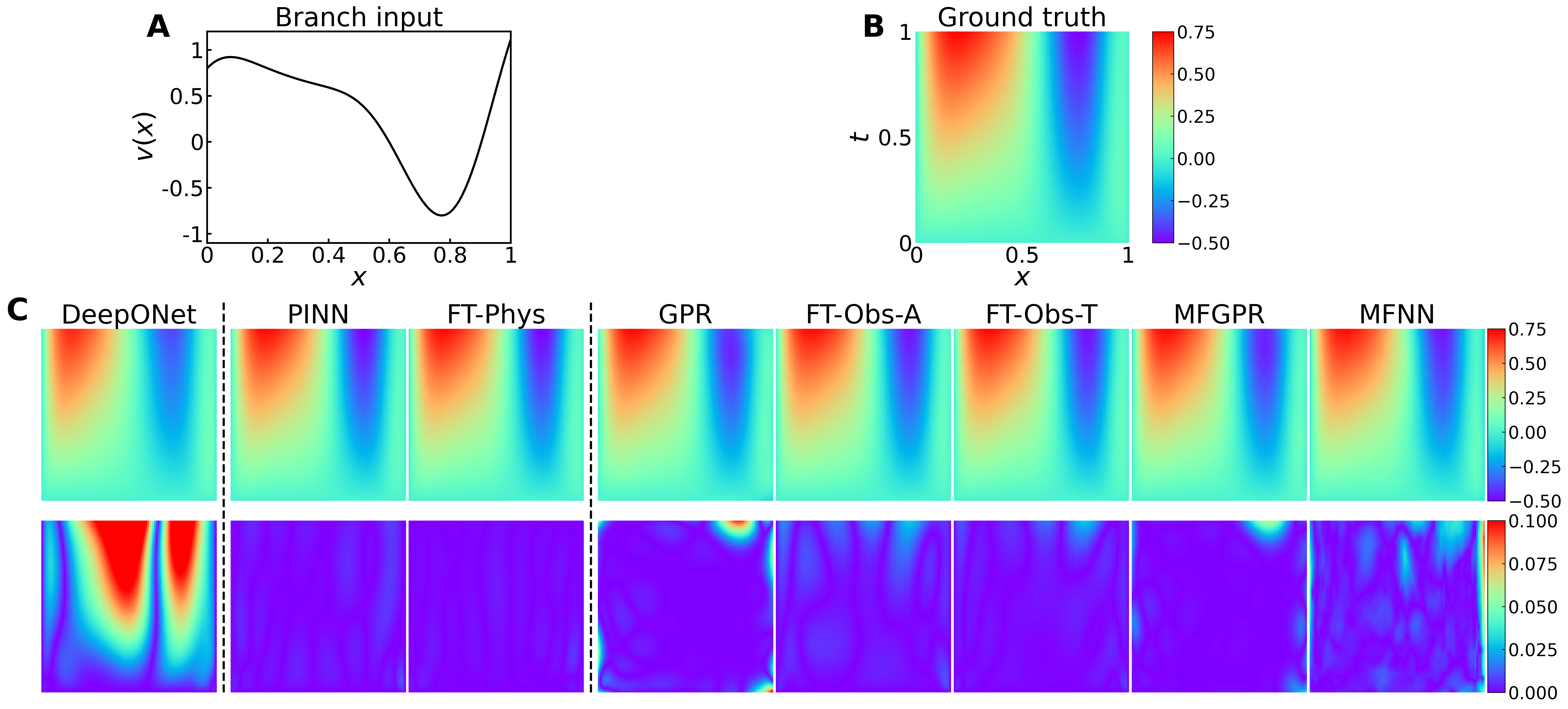}
     \caption{\textbf{An example of extrapolation for diffusion-reaction equation in Section~\ref{subsec:diffusion-reaction}.} (\textbf{A}) A test input function. (\textbf{B}) The corresponding PDE solution. (\textbf{C}) Predictions (first row) and errors (second row) of different methods. FT-Phys and FT-Obs-T obtain the best results.}
     \label{fig:dr-heatmap}
\end{figure}

Fig.~\ref{fig:dr-heatmap} is an example illustrating the predictions and absolute errors of all methods when we have 100 observations. We find that the locations of significant errors are similar between GPR and MFGPR (Fig.~\ref{fig:dr-heatmap}C) due to the similarity of these two methods. FT-Obs-A and FT-Obs-T also have a similar profile of errors. It is worth noting that both PINN and FT-Phys have very low errors, so we cannot find a similar error between PINN and FT-Phys in this example, but it can be found in the advection equation in Section~\ref{subsec:advection}.

\paragraph{Detailed results of FT-Phys.}
We consider physics as additional information and the approach of fine-tuning with physics. Seven different learning rates (i.e., 0.01, 0.005, 0.002, 0.001, 0.0005, 0.0002, and 0.0001) are used to fine-tune the pre-trained DeepONet. Appendix Figs.~\ref{fig:diff-reac}A--G display the $L^2$ relative errors of 100 test functions for different learning rates. When the number of iterations is small, PINNs give larger $L^2$  relative errors ($>$ 100\%), while the FT-Phys produces much lower errors ($\sim$10\%). This is because the parameters of PINNs are randomly initialized. By contrast, the initial parameters of the FT-Phys are given by the pre-trained DeepONet. For fine-tuning different parts of DeepONet, we select the best accuracy among different learning rates and summarize the results in Appendix Fig.~\ref{fig:diff-reac}H. Appendix Fig.~\ref{fig:diff-reac}I shows the $L^2$ relative errors with respect to the learning rate for different approaches. The performance of PINN is susceptible to the change in learning rates (Appendix Fig.~\ref{fig:diff-reac}I, black line). Fine-tuning with branch net ($\sim$1\%) in general performs worse than other approaches regardless of learning rates. Fine-tuning with the entire DeepONet (branch \& trunk) or with the trunk net performs the best and reaches a very low error for any learning rate between 0.01 and 0.0001. Fine-tuning with the last layer of the trunk net also achieves good accuracy. The best accuracy of different FT-Phys approaches is shown in Table~\ref{table:diffusion-reaction}.

\paragraph{Effect of $\lambda$ on FT-Obs-T.}
We aim to find an optimal $\lambda$ to achieve the lowest test errors. Also, we determine the effect of $\lambda$ on test errors when using different numbers of observed points and different $l_{\text{test}}$. The results of 12 cases with different new observation numbers and testing correlation lengths are shown in Appendix Fig.~\ref{fig:k}. We can see that the lowest error is obtained when $\lambda$ is $\sim$0.3, so we choose $\lambda=0.3$ as the default value in our experiments.

\subsection{Burgers' equation}
\label{subsec:burgers}

Then, we consider the Burgers' equation
\begin{equation*}
    \frac{\partial u}{\partial t} + u \frac{\partial u}{\partial x} = \nu \frac{\partial^2 u}{\partial x^2}, \quad x \in [0,1], \quad t \in [0,1],
\end{equation*}
with a periodic boundary condition and an initial condition $u_0(x)=v(x)$. In this study, $\nu$ is set at 0.1. Our goal is to train a DeepONet to learn the operator mapping from initial condition $v(x)$ to the solution $u(x, t)$. The periodic function $v(x)$ is sampled from a GRF with an exponential sine squared kernel (or periodic kernel) given by
$$k_l(x_1, x_2)=\exp\left(-\frac{2\sin^2(\pi||x_1-x_2||/p)}{l^2}\right),$$
where $l$ is the correlation length of the kernel, and $p$ is the periodicity of the kernel, which is chosen at 1. The correlation length for training is $l_{\text{train}}=1.0$, and to test the Ex.\textsuperscript{+}, we generate a test dataset of 100 functions with $l_{\text{test}}=0.6$.

\begin{figure}[htbp]
    \centering
    \includegraphics[width=\textwidth]{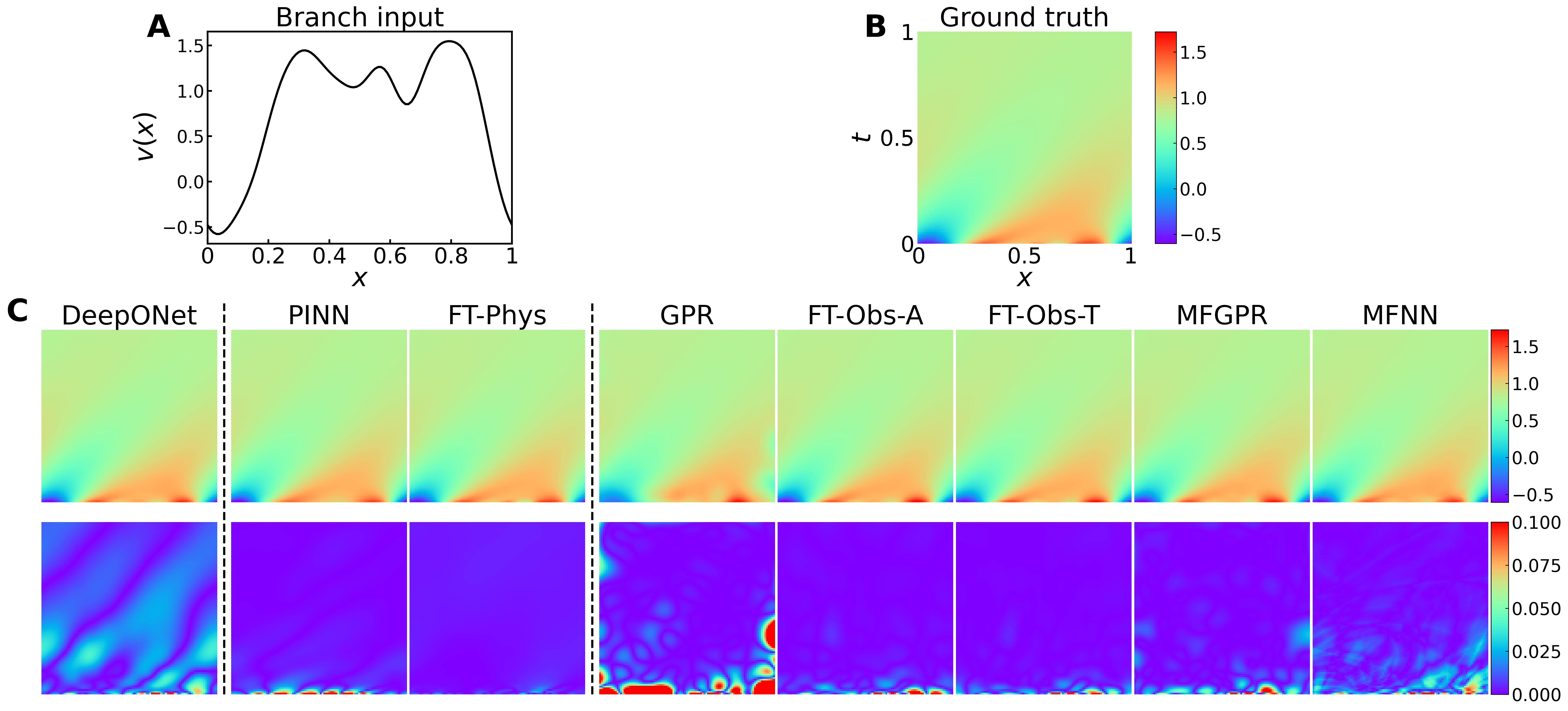}
     \caption{\textbf{An example of extrapolation for the Burgers' equation in Section~\ref{subsec:burgers}.} (\textbf{A}) A test input function. (\textbf{B}) The corresponding PDE solution. (\textbf{C}) Predictions (first row) and errors (second row) of different methods. FT-Phys and FT-Obs-T obtain the best results.}
     \label{fig:burgers-heatmap}
\end{figure}

Table~\ref{table:burgers} summarizes the $L^2$ relative errors of different methods. The pre-trained DeepONet has an average $L^2$ relative error of 6.53\%, and PIDeepONet has an average $L_2$ relative error of 9.27\% for the test dataset. As shown in Table~\ref{table:burgers}, when we have physics, fine-tuning with the trunk net at a learning rate of 0.002, or with the branch \& trunk nets at a learning rate of 0.001 can achieve low $L^2$ relative errors (1.42\%) and significantly outperform the pre-trained DeepONet. When we have new observations, unlike the results of the diffusion-reaction equation, FT-Obs-A and FT-Obs-T consistently outperform MFGPR and MFNN, while FT-Obs-T works the best among methods that use new observations. Fig.~\ref{fig:burgers-heatmap} is an example of illustrating the prediction and absolute errors of all methods when we have 200 observations. In this example, pre-trained DeepONet has relatively large errors distributed in the whole domain. In contrast, the large errors of the remaining methods are gathered in the initial area of the domain.

\begin{table}[htbp]
    \centering
    \caption{\textbf{$L^2$ relative error of different methods for the Burgers' equation in Section~\ref{subsec:burgers}.} $l_{\text{train}}=1.0$ and $l_{\text{test}}=0.6$. Bold font indicates the smallest two errors in each case, and the underlined text indicates the smallest error.}
    \label{table:burgers}
    \begin{tabular}{lcccc}
    \toprule
    DeepONet & \multicolumn{4}{c}{ Error(In.): $2.21\pm1.11\%$ \quad Error(Ex.\textsuperscript{+}): $6.53\pm3.33\%$} \\
    PIDeepONet & \multicolumn{4}{c}{ Error(In.): $4.96\pm1.80\%$ \quad Error(Ex.\textsuperscript{+}): $9.27\pm3.93\%$} \\
    \midrule
    PINN & \multicolumn{4}{c}{$2.79 \pm 1.27$\%} \\
    \hdashline
    & Branch \& Trunk & Branch & Trunk & Trunk last \\  
    FT-Phys &\underline{$\pmb{1.42}\pm\pmb{0.85}$}\%& $6.32\pm3.06$\% &  \underline{$\pmb{1.42}\pm\pmb{0.93}$}\%&$1.73\pm0.94$\% \\
    \midrule
    & 20 points & 50 points & 100 points & 200 points\\
    \hdashline
    GPR & $43.0 \pm 20.9 \%$ & $29.2 \pm 15.1 \%$ & $18.6\pm 10.3\%$  &  $12.8\pm 7.58\%$ \\
    \hdashline
    FT-Obs-A & $\pmb{4.97} \pm \pmb{2.57}\%$ & $ \pmb{4.40} \pm \pmb{2.27} \% $& $\pmb{4.00} \pm \pmb{1.99}\%$ & $\pmb{3.78} \pm \pmb{2.02} \%$\\
    FT-Obs-T & \underline{$\pmb{4.53} \pm \pmb{2.32}$}\% & \underline{$\pmb{3.89} \pm \pmb{1.98}$}\% & \underline{$\pmb{3.44} \pm \pmb{1.72}$}\% & \underline{$\pmb{3.07} \pm \pmb{1.55}$}\%\\
    MFGPR & $6.28\pm3.20 \%$ & $5.79\pm3.29 \%$ & $5.15\pm2.97 \%$ & $4.22\pm2.31\%$ \\
    MFNN & $6.34\pm3.23 \%$ & $5.62\pm 2.84 \%$ & $5.15\pm2.56\%$ & $4.59\pm2.33\%$\\    
    \bottomrule
    \end{tabular}
\end{table}

\paragraph{Detailed results of FT-Phys.}
We consider the approach of fine-tuning with physics. Seven different learning rates (i.e., 0.01, 0.005, 0.002, 0.001, 0.0005, 0.0002, and 0.0001) are used to fine-tune the pre-trained DeepONet. Appendix Figs.~\ref{fig:burgers}A--G display the $L^2$ relative errors of 100 test functions for different learning rates. When the number of iterations is small, PINNs give larger $L^2$  relative errors ($\sim$ 100\%), while the FT-Phys produces much lower errors ($\sim$7\%), which is consistent with the results of the diffusion-reaction equation. For fine-tuning different parts of DeepONet, we select the best accuracy among different learning rates and summarize the results in Appendix Fig.~\ref{fig:burgers}H. Appendix Fig.~\ref{fig:burgers}I shows the $L^2$ relative errors with respect to the learning rate for different approaches. The performance of PINN is very sensitive to the change in learning rates (Appendix Fig.~\ref{fig:burgers}I, black line). When the learning rate is low ($\sim$0.0001), the $L^2$ relative error of PINN is large ($\sim$20\%), while PINN can achieve a small $L^2$ relative error ($\sim$3\%) when the learning rate is 0.002. Fine-tuning with the branch net ($\sim$6\%) performs worse than other approaches regardless of learning rates. Fine-tuning with the entire DeepONet (branch \& trunk), with the trunk net, or with the last trunk layer performs similarly and reaches low error for any learning rate between 0.0001 and 0.002, while their $L^2$ relative errors increase with a large learning rate ($>$0.002).

\subsection{Advection equation}
\label{subsec:advection}

Next, we consider the advection equation
\begin{equation*}
    \frac{\partial u}{\partial t} + v(x)\frac{\partial u}{\partial x} = 0, \quad x \in [0,1], \quad t \in [0,1],
\end{equation*}
with the initial condition $u(x, 0)=\sin(\pi x)$ and boundary condition $u(0, t)=\sin(\pi t/2)$. To make sure parameter $v(x)$ is larger than 0, we take it in the form of $v(x)=V(x)-\min_xV(x)+1$. We train a DeepONet to learn the operator mapping from $v(x)$ to the solution $u(x, t)$. $V(x)$ is sampled from a GRF with an RBF kernel with the correlation length $l_{\text{train}}=0.5$. To test the Ex.\textsuperscript{+}, we generate a test dataset of 100 functions with $l_{\text{test}}=0.2$.

\begin{figure}[htbp]
    \centering
    \includegraphics[width=\textwidth]{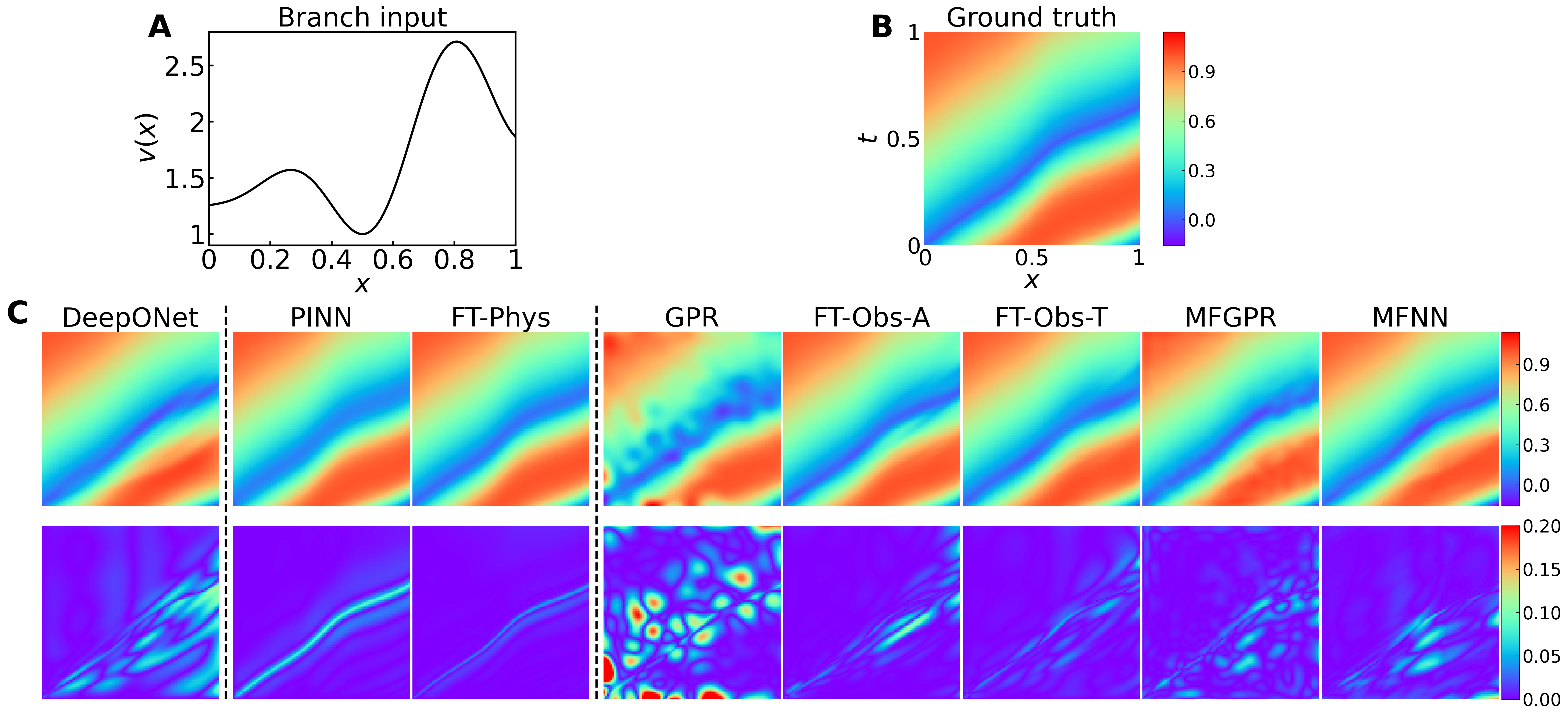}
     \caption{\textbf{An example of extrapolation for advection equation in Section~\ref{subsec:advection}.} (\textbf{A}) A test input function. (\textbf{B}) The corresponding PDE solution. (\textbf{C}) Predictions (first row) and errors (second row) of different methods. FT-Phys and FT-Obs-T obtain the best results.}
     \label{fig:advection-heatmap}
\end{figure}

Table~\ref{table:advection} summarizes the $L^2$ relative errors of different methods. The pre-trained DeepONet has an average $L^2$ relative error of 8.75\%, and PIDeepONet has an average $L_2$ relative error of 11.3\% for the test dataset. When we have physics, fine-tuning with the trunk net achieves low $L^2$ relative errors (0.93\%). When we have new observations, like the results of Burgers' equation, FT-Obs-A and FT-Obs-T always outperform MFGPR and MFNN, and FT-Obs-T works the best. Fig.~\ref{fig:advection-heatmap} is an example illustrating the prediction and absolute errors of all methods when we have 200 observations. In this example, PINN and FT-Phys have similar error profiles, and FT-Phys is more accurate than PINN, since the pre-trained DeepONet gives better initial parameters of the FT-Phys. Similar error profiles are also observed between FT-Obs-A and FT-Obs-T.

\begin{table}[htbp]
    \centering
    \caption{\textbf{$L^2$ relative error of different methods for the advection equation in Section~\ref{subsec:advection}.} $l_{\text{train}}=0.5$ and $l_{\text{test}}=0.2$.  Bold font indicates the smallest two errors in each case, and the underlined text indicates the smallest error.}
    \label{table:advection}
    \begin{tabular}{lcccc}
    \toprule
    DeepONet & \multicolumn{4}{c}{ Error(In.): $1.30\pm0.26\%$ \quad Error(Ex.\textsuperscript{+}): $8.75\pm6.42\%$} \\
    PIDeepONet & \multicolumn{4}{c}{ Error(In.): $3.60\pm0.70\%$ \quad Error(Ex.\textsuperscript{+}): $10.4\pm4.48\%$} \\
    \midrule
    \midrule
    PINN & \multicolumn{4}{c}{$1.67 \pm 0.53$\%} \\
    \hdashline
    &Branch \& Trunk& Branch&Trunk&Trunk last\\
    FT-Phys &$1.05\pm0.3$\%&$2.65\pm1.17$\%& \underline{$\pmb{0.93}\pm\pmb{0.23}$}\% &$\pmb{1.01}\pm\pmb{0.31}$\%\\     
    \midrule
    \midrule
    & 20 points & 50 points & 100 points & 200 points\\
    \hdashline
    GPR & $34.6 \pm 9.46 \%$ & $25.8 \pm 7.89 \%$ & $16.7\pm 3.99\%$  &  $11.6\pm 3.54\%$ \\
    \hdashline
    FT-Obs-A & $\pmb{6.15}\pm\pmb{2.91}\%$ & $\pmb{5.79}\pm\pmb{2.43} \% $& $\pmb{3.91}\pm\pmb{1.52}\%$ & $\pmb{2.42}\pm\pmb{0.79} \%$\\
    FT-Obs-T & \underline{\pmb{5.07} $\pm$ \pmb{2.34}}\% & \underline{\pmb{4.08} $\pm$ \pmb{1.73}}\% & \underline{\pmb{3.16} $\pm$ \pmb{1.34}}\% & \underline{\pmb{2.00} $\pm$ \pmb{0.74}}\% \\
    MFGPR & $8.39\pm5.38 \%$ & $6.80\pm4.00 \%$ & $5.47\pm2.87 \%$ & $4.19\pm2.01\%$ \\
    MFNN & $8.46\pm4.72 \%$ & $6.20\pm 2.81 \%$ & $6.90\pm6.47\%$ & $4.64\pm3.55\%$\\
    \bottomrule
    \end{tabular}
\end{table}

Moreover, we consider different Ex.\textsuperscript{+} level by using different correlation lengths for testing (Table~\ref{table:advection_l}). Test datasets of 100 functions are generated from GFR with $l_{\text{test}}=0.2$, 0.15, 0.1, and 0.05. We use 100 data points for fine-tuning methods. Decreasing the correlation lengths represents an increasing trend of extrapolation and results in larger error, which also validates the result in Section~\ref{sec:U-curve}. All proposed methods outperform the baseline models, DeepONet and GPR. When we have physics, FT-Phys outperforms PINN in all scenarios, and even under the great extrapolation with $l_{\text{test}}=0.05$, it reaches a satisfactory error as low as 3.22\%. With 100 measurements, FT-Obs-T achieves better performance compared with other fine-tune methods. For multifidelity methods, MFNN performs better as the level of extrapolation increases, but both are slightly worse than FT-Obs methods.

\begin{table}[htbp]
    \centering
    \caption{\textbf{$L^2$ relative error of different methods under different testing correlation lengths for the advection equation in Section~\ref{subsec:advection}.} Bold font indicates the smallest error in each case. $W_2$ is the 2-Wasserstein distance between the training space and testing space.} 
    \label{table:advection_l}
    \begin{tabular}{lcccc}
    \toprule
    $l_{\text{test}}$& 0.20 & 0.15 & 0.10 & 0.05 \\
    $W_2$&0.4578&0.5945&0.7606&0.9721\\
    \midrule  
    \midrule
    DeepONet & $8.75\pm6.42 \%$ & $14.5\pm10.1 \%$ & $19.3\pm12.2 \%$ & $25.6\pm14.9 \%$ \\
    PIDeepONet & $10.4\pm4.48 \%$ & $14.4\pm6.26 \%$ & $20.4\pm8.53 \%$ & $27.8\pm9.08 \%$ \\
    \midrule
    \midrule
    PINN & $1.67\pm0.53 \%$ & $1.78\pm0.51 \%$ & $2.27\pm0.71 \%$ & $3.57\pm1.07 \%$  \\
    \hdashline
    FT-Phys & $\pmb{0.93}\pm\pmb{0.23} \%$ & $\pmb{1.05}\pm\pmb{0.28} \%$ & $\pmb{1.53}\pm\pmb{0.43} \%$ & $\pmb{3.22}\pm\pmb{1.16} \%$  \\     
    \midrule
    \midrule
    GPR & $16.7\pm3.99 \%$ & $16.8\pm4.55 \%$ & $18.0\pm4.63 \%$ & $18.4\pm5.37 \%$  \\
    \hdashline
    FT-Obs-A & $3.91\pm1.52 \%$ & $6.61\pm2.66 \%$ & $6.61\pm2.34 \%$ & $10.8\pm4.40 \%$ \\
    FT-Obs-T & $\pmb{3.16}\pm\pmb{1.34} \%$ & $\pmb{4.65}\pm\pmb{1.89} \%$ & $\pmb{6.07}\pm\pmb{2.23} \%$ & $\pmb{7.21}\pm\pmb{2.39} \%$ \\
    MFGPR & $5.47\pm2.87 \%$ & $8.61\pm3.94 \%$ & $10.9\pm3.92 \%$ & $13.4\pm4.14 \%$ \\
    MFNN & $6.90\pm6.47 \%$ & $8.05\pm5.05 \%$ & $8.58\pm4.22 \%$ & $10.6\pm9.16 \%$ \\
    \bottomrule
    \end{tabular}
\end{table}

\paragraph{Detailed results of FT-Phys.}
We consider the approach of fine-tuning with physics. Seven different learning rates (i.e., 0.01, 0.005, 0.002, 0.001, 0.0005, 0.0002, and 0.0001) are used to fine-tune the pre-trained DeepONet. Appendix Figs.~\ref{fig:advection}A--G display the $L^2$ relative errors of 100 test functions for different learning rates. When the number of iterations is small, PINNs give larger $L^2$ relative errors ($>$100\%). In comparison, the FT-Phys produces much lower errors ($\sim$10\%), which is consistent with the results of the diffusion-reaction equation and Burgers' equation. For fine-tuning different parts of DeepONet, we select the best accuracy among different learning rates and summarize the results in Appendix Fig.~\ref{fig:advection}H. Appendix Fig.~\ref{fig:advection}I shows the $L^2$ relative errors with respect to learning rate for different approaches. The performance of PINN is susceptible to the change in learning rates (Appendix Fig.~\ref{fig:advection}I, black line). When the learning rate is low ($\sim$0.0001), the $L^2$ relative error of PINN is large ($\sim$10\%), while PINN can achieve a small $L^2$ relative error ($\sim$2\%) when the learning rate is 0.005. Fine-tuning with branch net ($\sim$3\%) performs worse than other approaches regardless of learning rates. Fine-tuning with the entire DeepONet (branch \& trunk), with the trunk net, or with the last trunk layer performs similarly and reaches a low error($\sim$1\%) at any learning rate between 0.01 and 0.0001.

\subsection{Poisson equation in a triangular domain with notch}
\label{subsec:poisson}

To evaluate the performance of our methods in an unstructured domain, we consider the Poisson equation in a triangular domain with a notch
\begin{equation*}
    \frac{\partial^2 u}{\partial x^2} + \frac{\partial^2 u}{\partial y^2} + 10 = 0, \quad (x, y) \in \Omega,
\end{equation*}
where $\Omega$ is a concave heptagon whose vertices are [0, 0], [0.5, 0.866], [1, 0], [0.51, 0], [0.51, 0.4], [0.49, 0.4], [0.49, 0]. The boundary condition $u(x, y)|_{\partial\Omega}=v(x)$, is only the function its $x$ coordinate. We train a DeepONet to learn the operator mapping from boundary condition $v(x)$ to the solution $u(x, t)$. $v(x)$ is sampled from a GRF with an RBF kernel with the correlation length $l_{\text{train}}=0.5$. To test the Ex.\textsuperscript{+}, we generate a test dataset of 100 functions with $l_{\text{test}}=0.2$. The reference solution is obtained by the PDEtoolbox in Matlab with an unstructured mesh of 5082 nodes.

Table~\ref{table:poisson} summarizes the $L^2$ relative errors of different methods. The pre-trained DeepONet has an average $L^2$ relative error of 14.8\% for the test dataset. In this problem, we only tested the case of new observations, as FT-Phys and PINN failed to achieve a good accuracy due to the complex domain geometry with singularity points. Like the results of Burgers' equation and advection equation, FT-Obs-A and FT-Obs-T always outperform MFGPR and MFNN. The FT-Obs-T has the lowest errors (0.95\% with 200 observations) among all the methods.

\begin{table}[htbp]
    \centering
    \caption{\textbf{$L^2$ relative error of different methods for the Poisson equation in Section~\ref{subsec:poisson}.} $l_{\text{train}}=0.5$ and $l_{\text{test}}=0.2$.  Bold font indicates the smallest two errors in each case, and the underlined text indicates the smallest error.}
    \label{table:poisson}
    \begin{tabular}{lcccc}
    \toprule
    DeepONet & \multicolumn{4}{c}{ Error(In.): $0.09\pm0.04\%$ \quad Error(Ex.\textsuperscript{+}): $14.8\pm12.0\%$} \\ 
    \midrule
    \midrule
    & 20 points & 50 points & 100 points & 200 points\\
    \hdashline
    GPR & $16.7 \pm 9.51 \%$ & $12.1 \pm 6.57 \%$ & $8.06\pm 4.99\%$  &  $6.26\pm 4.63\%$ \\
    \hdashline
    FT-Obs-A & $\pmb{7.52}\pm\pmb{5.57}\%$ & $\pmb{4.15}\pm\pmb{2.89} \% $& $\pmb{2.49} \pm \pmb{1.64}\%$ & $\pmb{1.80} \pm \pmb{1.06}\%$\\
    FT-Obs-T & \underline{\pmb{4.97} $\pm$ \pmb{3.14}}\% & \underline{\pmb{3.01} $\pm$ \pmb{2.11}}\% & \underline{\pmb{1.55} $\pm$ \pmb{1.08}}\% & \underline{\pmb{0.95} $\pm$ \pmb{0.56}}\%\\
    MFGPR & $8.00\pm5.58 \%$ & $5.25\pm3.38 \%$ & $3.38\pm1.97 \%$ & $2.42\pm1.39\%$ \\
    MFNN & $11.32\pm6.96 \%$ & $7.63\pm5.16 \%$ & $4.64\pm2.75\%$ & $3.12\pm1.51\%$\\
    \bottomrule
    \end{tabular}
\end{table}

\begin{figure}[htbp]
    \centering
    \includegraphics[width=\textwidth]{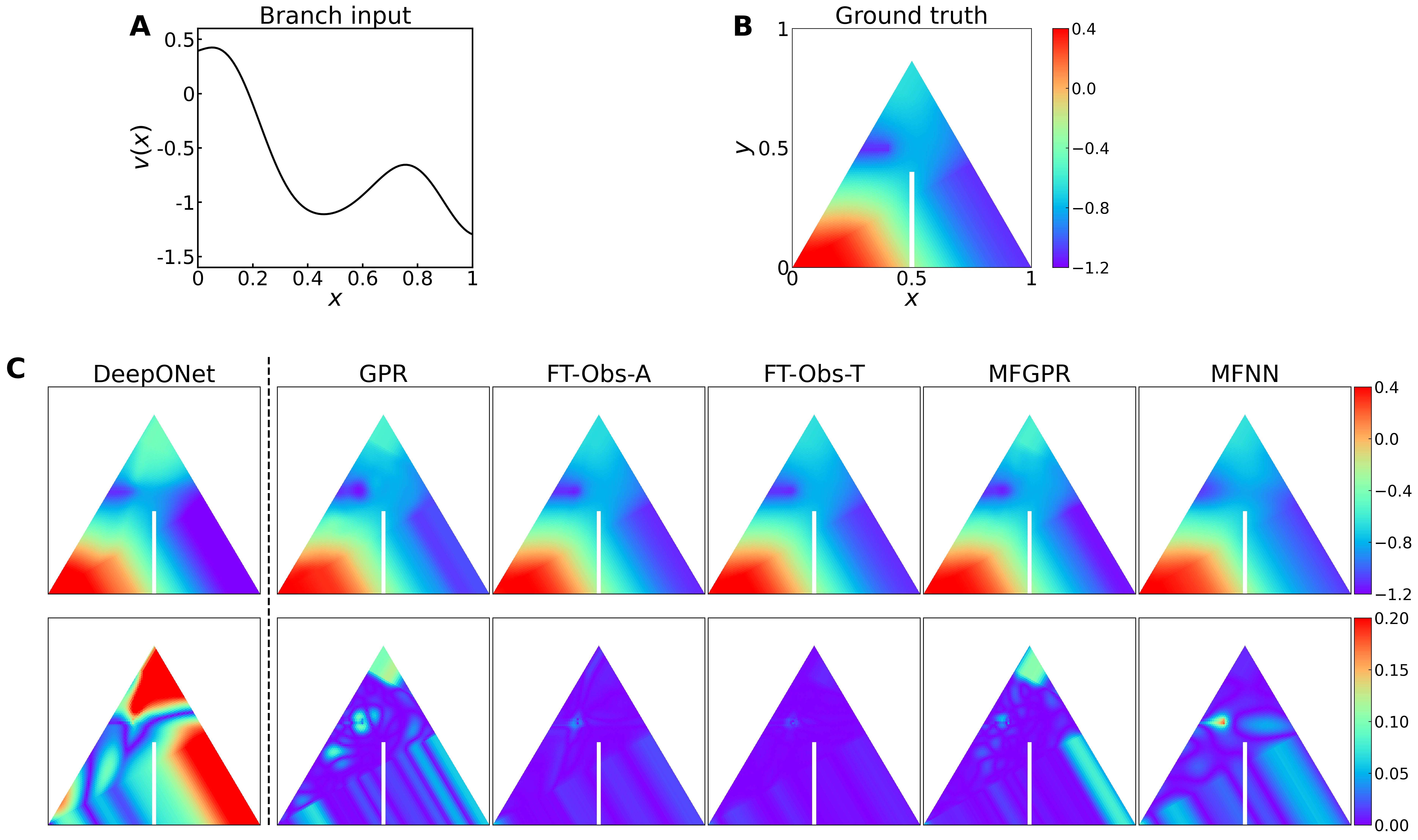}
     \caption{\textbf{An example of extrapolation for the Poisson equation in a triangular domain with a notch in Section~\ref{subsec:poisson}.} (\textbf{A}) A test input function. (\textbf{B}) The corresponding PDE solution. (\textbf{C}) Predictions (first row) and errors (second row) of different methods. FT-Obs-T obtains the best result.}
     \label{fig:poisson-heatmap}
\end{figure}

Fig.~\ref{fig:poisson-heatmap} is an example of illustrating the prediction and absolute errors of all methods when we have 200 observations. Similar to the results in Section~\ref{subsec:diffusion-reaction}, GPR and MFGPR have similar error patterns, and MFGPR is more accurate than GPR due to the additional low-fidelity dataset. FT-Obs-T and FT-Obs-A outperform other methods, and FT-Obs-T is slightly better than FT-Obs-A.

\begin{figure}[htbp]
    \centering
    \includegraphics[width=0.6\textwidth]{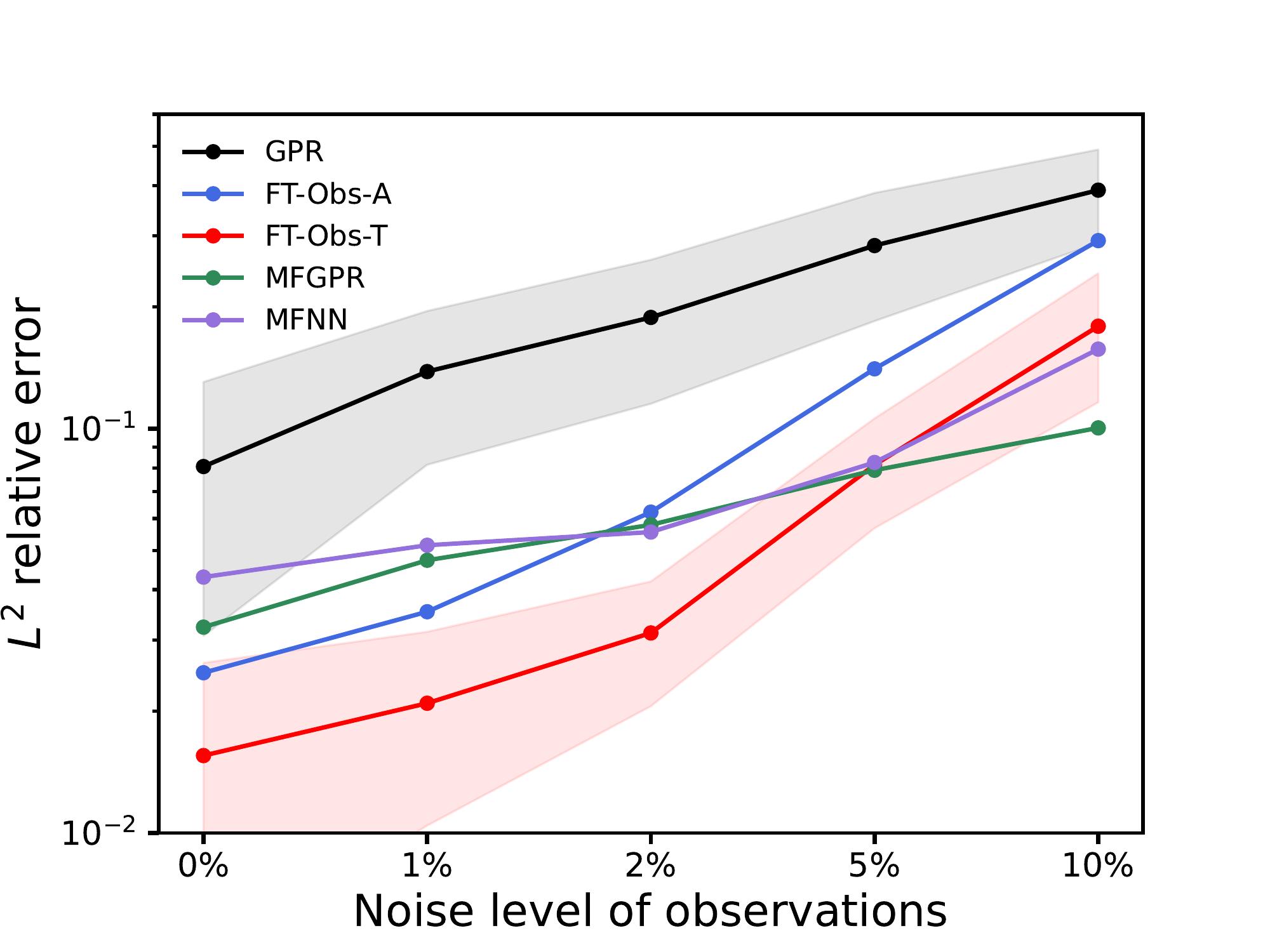}
     \caption{\textbf{$L^2$ relative error for observations with different noise level for the Poisson equation in Section~\ref{subsec:poisson}.}}
     \label{fig:poisson-noise}
\end{figure}

To validate the robustness of proposed fine-tuning methods, we test the performance under different levels of noise in the observations (Fig.~\ref{fig:poisson-noise}). With other settings remaining unchanged, increasing noise levels lead to a corresponding increase in $L_2$ relative error. All fine-tuning methods exhibit better performance than GPR. When the noise level is less than 5\%, FT-Obs-T reaches the best accuracy and as the noise gradually increases to 10\%, MFGPR method outperforms all other methods. Multifidelity methods are relatively insensitive to noise compared with FT-Obs.

\subsection{Lid-driven cavity flow in complex geometries}
\label{subsec:lid}

\begin{figure}[htbp]
    \centering
    \includegraphics[width=\textwidth]{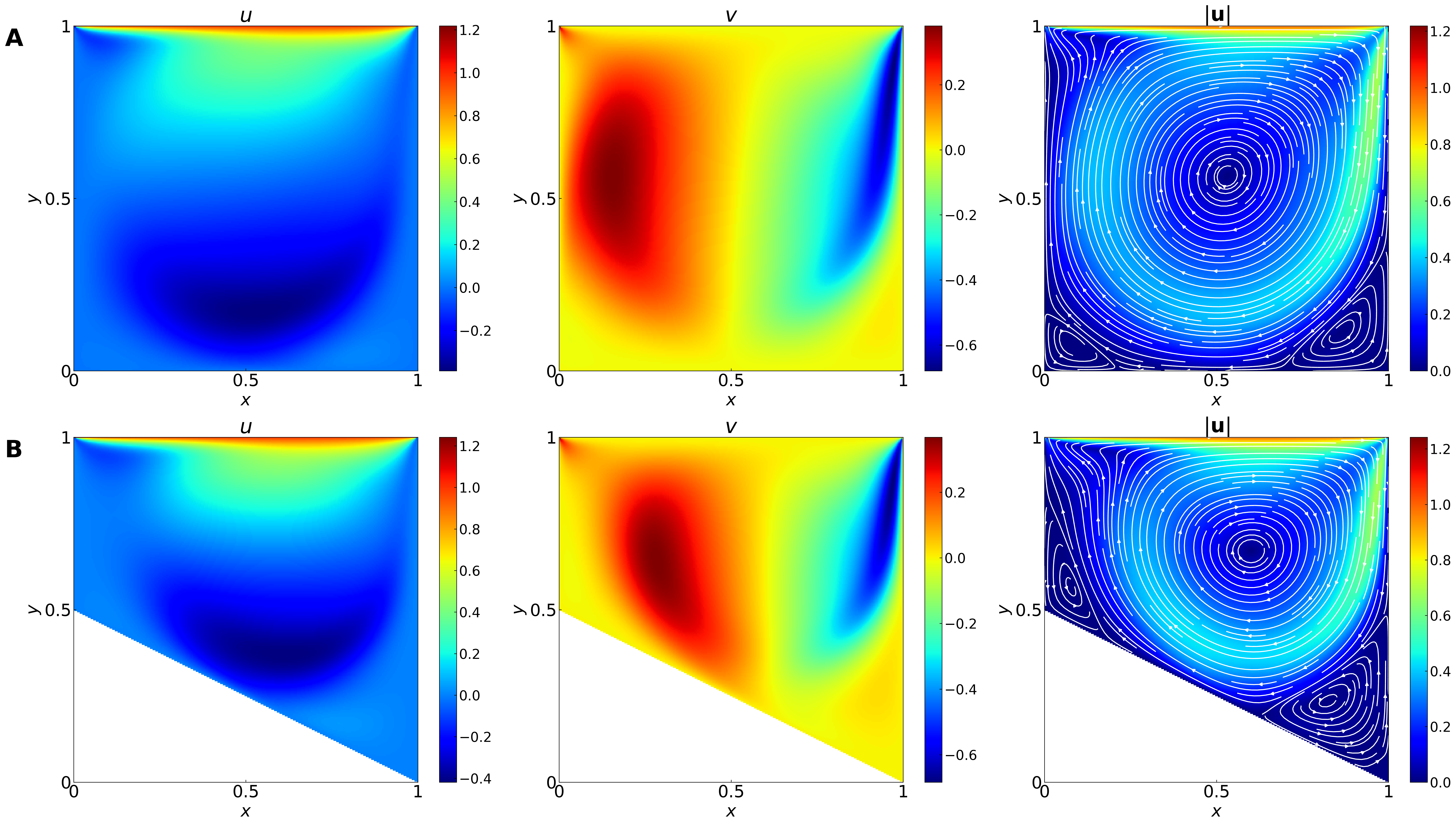}
     \caption{\textbf{Lid-driven cavity flows in two geometries in Section~\ref{subsec:lid}.} (\textbf{A}) $m=0$. (\textbf{B}) $m=0.5$. Re = 1000.}
     \label{fig:cavity}
\end{figure}

We evaluate the performance in the lid-driven cavity flow problem, which is a benchmark problem for viscous incompressible fluid flow. The incompressible flow is described by the Navier-Stokes equations,
\begin{gather*}
\frac{\partial \mathbf{u}}{\partial t} + (\mathbf{u} \cdot \nabla) \mathbf{u} = \nu \nabla^2 \mathbf{u} - \nabla p, \\
\nabla \cdot \mathbf{u}=0,
\end{gather*}
where $\mathbf{u}(\mathbf{x}, t) = (u, v)$ is the velocity field, $\nu$ is the kinematic viscosity, and $p(\mathbf{x}, t)$ is the pressure. The Reynolds number (Re) is chosen to be 1000. We consider different geometries by starting with a square with side length $l=1$ and gradually lifting the left bottom point with other three points fixed, i.e., the bottom line is described by $l(x) = -mx+m$ with $0\le m \le 0.5$. The top wall has a unit velocity in the $x$-direction. There is a flow circulation in the cavity, and increasing value of $m$ represents more extreme cases, see two examples in Fig.~\ref{fig:cavity}.

The goal is to learn the operator mapping from the boundary line $l(x)$ to the velocity $\mathbf{u}(\mathbf{x}, t)$. We take $m=0,0.02,0.04,\dots,0.4$ for generating the training dataset and select $m=0.41, 0.42, 0.43,\dots, 0.5$ for extrapolation testing. Larger $m$ represents more aggressive extrapolation. The reference solution is obtained by the finite element method with a nonuniform mesh of 10201 nodes. For fine-tuning with new observations and multifidelity methods, 100 points are randomly chosen as the new information. To avoid the randomness, this process is repeated for 10 times.

\begin{figure}[htbp]
    \centering
    \includegraphics[width=\textwidth]{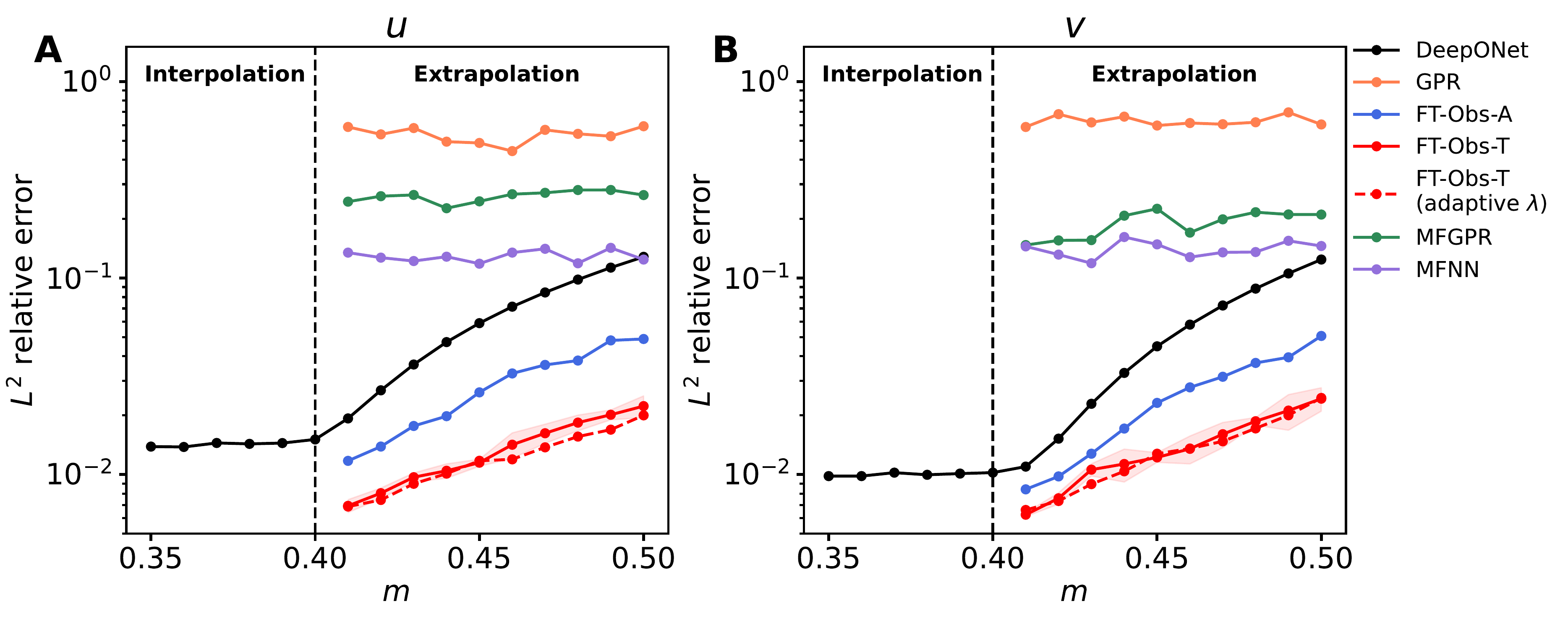}
    \caption{\textbf{$L^2$ relative error of different methods for the lid-driven cavity flow in Section~\ref{subsec:lid}.} (\textbf{A}) The $x$-component of velocity. (\textbf{B}) The $y$-component of velocity. The left part ($m \leq 0.40$) is interpolation, while the right part ($m>0.40$) is extrapolation.}
    \label{fig:cavity_result}
\end{figure}

For the prediction of the horizontal velocity $u$ (Fig.~\ref{fig:cavity_result}A), the baseline model GPR has the largest $L^2$ relative error. Multifidelity methods outperform the single-fidelity GPR, and MFNN has a better accuracy than MFGPR (Fig.~\ref{fig:cavity_result}A), yet both are still far from being satisfactory. We note that in previous examples, the RBF kernel works well for MFGPR, but in this problem, MFGPR with the RBF kernel has a large error, while the Matern kernel with $\nu=1.5$ performs better (Appendix Fig.~\ref{fig:cavity_mfgp}). Fine-tuning with observations provides a considerable improvement on the accuracy and reduces the $L_2$ relative error to $10^{-2}$. FT-Obs-T works the best among all proposed method. The prediction of the vertical velocity $v$ (Fig.~\ref{fig:cavity_result}B) also has a similar behavior.

In FT-Obs-T, besides taking the weight $\lambda$ as constant value, we also consider an approach to adaptively update the value of $\lambda$ during training. Specifically, we use $k = 0.1$ as the initial value, and then after every 100 iterations, $\lambda$ is updated by gradient descent, i.e.,
$$\lambda \leftarrow \lambda + \gamma_\lambda \frac{\partial \mathcal{L}_{\mathcal{T},\text{obs}}}{\partial \lambda},$$
where $\gamma_\lambda = 0.3$ is the learning rate. In this way, we increase the weight of the new information gradually during training. Introducing the adaptive adjustment of $\lambda$ further slightly improves the accuracy (Fig.~\ref{fig:cavity_result}).

\begin{figure}[htbp]
    \centering
    \includegraphics[width=\textwidth]{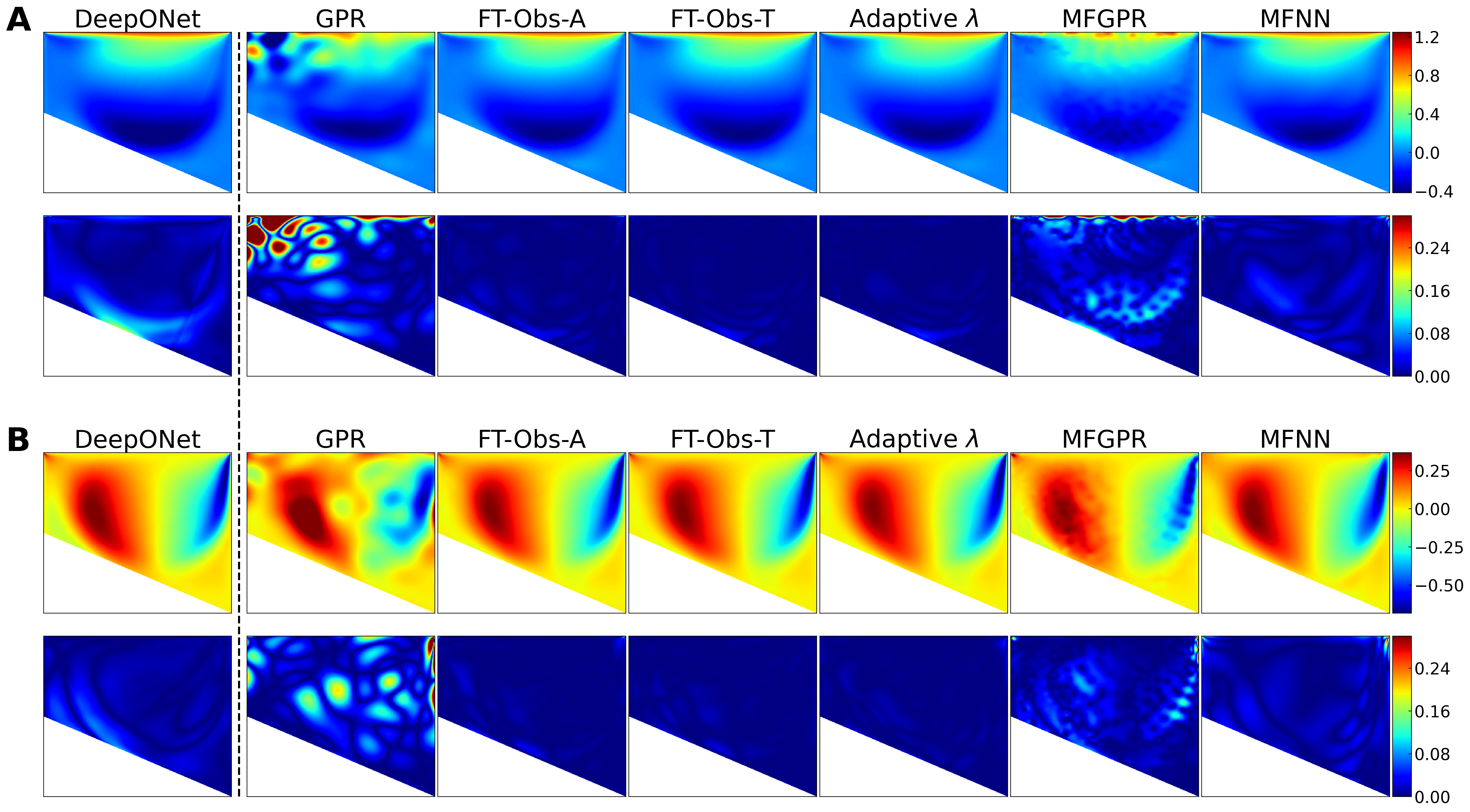}
     \caption{\textbf{An example of extrapolation ($m = 0.5$) for the lid-driven cavity flow in Section~\ref{subsec:lid}.} (\textbf{A}) Predictions (first row) and errors (second row) for the $x$-component of velocity. (\textbf{B}) Predictions (first row) and errors (second row) for the $y$-component of velocity. FT-Obs-T with adaptive weight (adaptive $\lambda$) obtain the best result.}
     \label{fig:cavity_heatmap}
\end{figure}

We show an example for illustrating the prediction and absolute errors of all methods when $m=0.5$ in Fig.~\ref{fig:cavity_heatmap}. In this example, most proposed models exhibit better predictions of the velocity field than pre-trained DeepONet and GPR. Among these methods, fine-tuning with observations, either alone (FT-Obs-A) or together (FT-Obs-T), have particularly accurate prediction of the velocity field in both $x$- and $y$-component. However, multifidelity methods have larger error at the region near central vortex and moving lid in the $x$-direction. This results from the limitation that finite number of data points may perform weakly or even fail in fitting these locations with relatively large gradient.

\begin{figure}[htbp]
    \centering
    \includegraphics[width=\textwidth]{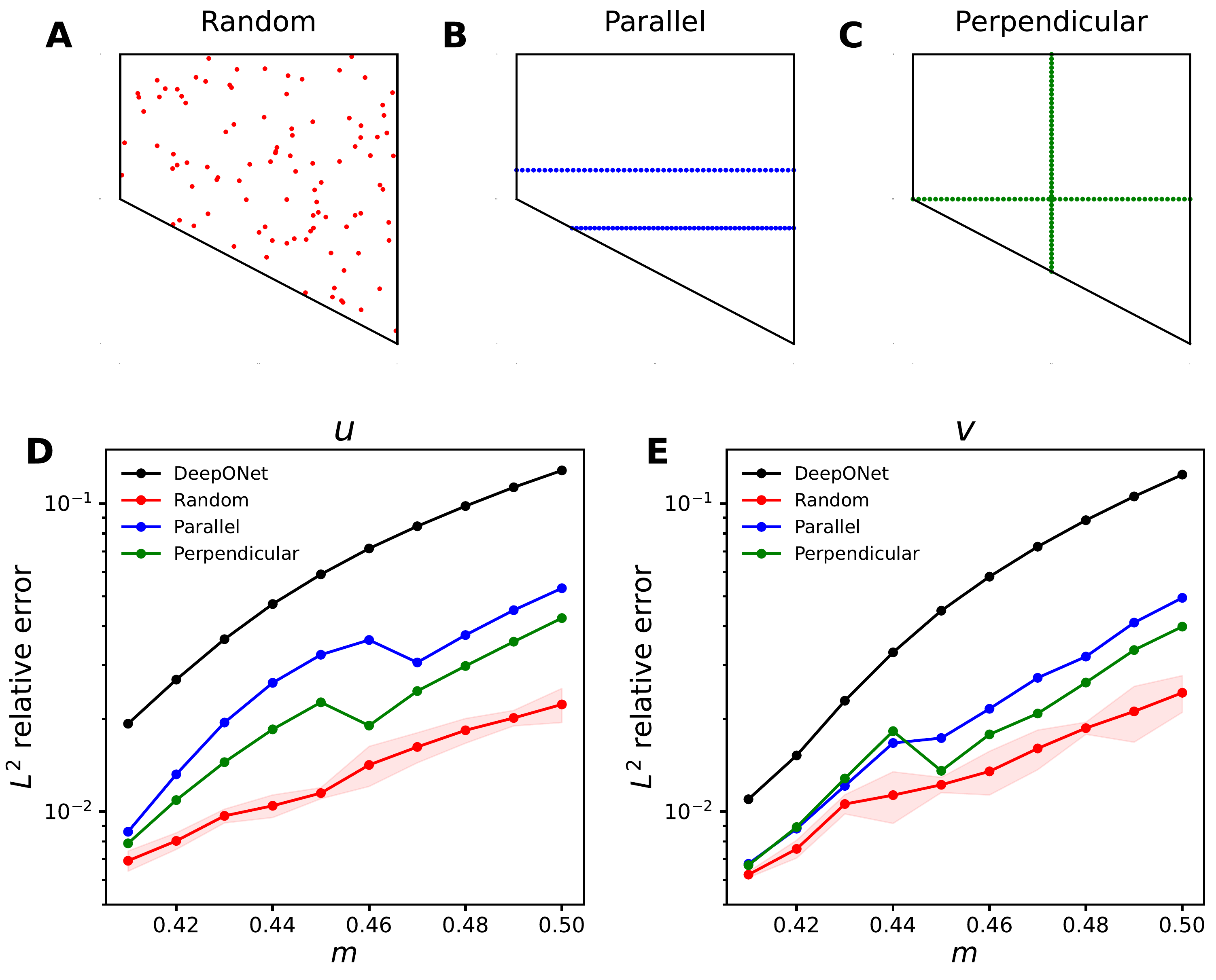}
    \caption{\textbf{FT-Obs-T for different data observations cases.} (\textbf{A}) Random sampling in the domain. (\textbf{B}) Sampling in two parallel lines. (\textbf{C}) Sampling in two perpendicular lines. (\textbf{D}) Extrapolation error of the $x$-component of velocity. (\textbf{E}) Extrapolation error of the $y$-component of velocity.}
    \label{fig:cavity_two_lines}
\end{figure}

For the experiments above, the new observations are randomly sampled in the domain, which might not be possible in practice, e.g. in experiments  (Fig.~\ref{fig:cavity_two_lines}A). Hence, we also consider two more realistic cases, where 50 data observations are uniformly sampled in certain lines (Figs.~\ref{fig:cavity_two_lines}B and C). For the ``Parallel'' case (Fig.~\ref{fig:cavity_two_lines}B), the new data points come from two horizontal lines, $y=0.4$ and $y=0.6$. For the ``Perpendicular'' case (Fig.~\ref{fig:cavity_two_lines}C), the new data points come from a horizontal line and a vertical line, $y=0.5$ and $x=0.5$. We use FT-Obs-T method for the three cases, each with 100 new observations. Random sampling in the domain leads to the most accurate prediction, while the other two sampling methods also achieve errors smaller than $<5\%$ (Figs.~\ref{fig:cavity_two_lines}D and E).

\section{Conclusions}
\label{sec:conclusion}

Having a new input in the Ex.\textsuperscript{+} region is inevitable in real-world applications and would lead to large errors and failure of NNs. In this study, we first present a systematic study of extrapolation error of deep operator networks (DeepONets). We provide a quantitative definition of the extrapolation complexity by 2-Wasserstein distance between two function spaces. Similar to interpolation error, we found a U-shaped error curve for extrapolation with respect to model capacity, such as network sizes and training iterations, but compared with interpolation and Ex.\textsuperscript{$-$}, the Ex.\textsuperscript{+} curve has larger error and earlier transition point. We also found that a larger training dataset is helpful for both interpolation and extrapolation.

To improve the prediction accuracy under Ex.\textsuperscript{+} scenarios, we consider additional information of physics or sparse observations. As the first step of the prediction workflow, we determine if the new input is in the region of interpolation or extrapolation. Given the governing partial differential equations (PDEs) of the system, we employ the PDE loss to fine-tune a pre-trained DeepONet (either the entire DeepONet or a part of the network). When we have extra sparse observations, we propose to either fine-tune a pre-trained DeepONet or apply a multifidelity learning framework. We demonstrate the excellent extrapolation capability of the proposed methods for diverse PDE problems. Furthermore, we validate the robustness of proposed methods by testing with different levels of noise.

We provide a practical guideline in choosing a proper extrapolation method depending on the available information, and desired accuracy and inference speed in Table~\ref{tab:conclusion}. When we have the physics as new information, fine-tuning with physics (FT-Phys) can achieve very high accuracy, and the computational cost of fine-tuning depends on the complexity of the PDEs. We found that only fine-tuning the trunk net usually works the best. When we have sparse observations, fine-tuning with observations (FT-Obs-A and FT-Obs-T) usually has better accuracy than multifidelity learning methods (MFGPR and MFNN). As FT-Obs-A only takes a small number of new observations, FT-Obs-A has a faster inference speed than FT-Obs-T, but FT-Obs-A usually has lower accuracy than FT-Obs-T due to overfitting and catastrophic forgetting. As MFGPR and MFNN need to train a new model from scratch, they have a relatively slow inference speed.

\begin{table}[htbp]
    \centering
    \caption{\textbf{Comparison of extrapolation accuracy and inference speed for different methods.} More stars ($\star$) represent better accuracy and faster speed.}
    \label{tab:conclusion}
    \begin{tabular}{l|l|cc}
        \toprule
         New information & Methods & Extrapolation accuracy & Inference speed \\
         \midrule
         Physics & FT-Phys (Trunk) & $\star \star \star$ & $\star \star$ \\
         \hdashline
         \multirow{4}{*}{Sparse observations}
         &FT-Obs-A & $\star \star$ & $\star\star\star$ \\
         &FT-Obs-T & $\star \star \star$ & $\star\star$ \\
         &MFGPR & $\star$ & $\star\star$\\
         &MFNN & $\star \star$ & $\star$ \\
         \bottomrule
    \end{tabular}
\end{table}

This study is the first attempt to understand and address extrapolation of deep neural operators, and more work should be done both theoretically and computationally. We observed the U-shaped error curve for extrapolation, but there could exist a double-descent behavior when the model capacity is further increased, which will be investigated in future work. As we show in FT-Phys, fine-tuning the trunk net or the entire DeepONet has better accuracy than fine-tuning the branch net. Similar studies for FT-Obs-A and Ft-Obs-T should also be conducted. In this study, we consider either the complete physics or sparse observations as the new information, but in practice, we may have sparse observations with partial physics simultaneously, and a corresponding efficient method should be developed. On the theoretical side, there have been a few efforts to address interpolation errors for specific forms of the operator~\cite{lanthaler2022error,deng2022approximation,de2022generic,kovachki2021universal,marcati2021exponential,herrmann2022neural,schwab2022deep}. A theoretical understanding of extrapolation error could be even more challenging.

\section*{Acknowledgments}

This work was supported by the U.S. Department of Energy [DE-SC0022953] and OSD/AFOSR MURI, USA grant FA9550-20-1-0358.

\appendix
\section{Abbreviations and notations}
\label{sec:notation}

We list in Table~\ref{tab:notation} the main abbreviations and notations that are used throughout this paper.

\begin{table}[htbp]
    \centering
    \caption{\textbf{Abbreviations and notations.}}
    \label{tab:notation}
    \begin{tabular}{ll}
        \toprule
        In. & Interpolation\\
        Ex.\textsuperscript{-} & Extrapolation when $l_{\text{train}}<l_{\text{test}}$\\
        Ex.\textsuperscript{+} & Extrapolation when $l_{\text{train}}>l_{\text{test}}$\\
        FT-Phys & Fine-tune with physics in Section~\ref{sec:physics}\\
        FT-Obs-A & Fine-tune with new observations alone in Section~\ref{sec:fine-tune}\\
        FT-Obs-T & Fine-tune with training data and new observations together in Section~\ref{sec:fine-tune}\\
        MFNN & Multifidelity neural networks in Section~\ref{sec:mf}\\
        MFGPR & Multifidelity Gaussian process regression in Section~\ref{sec:mf}\\
        \midrule
        $l$ & correlation length of Gaussian random field\\
		$\mathcal{G}$ & operator to learn\\
		$\mathcal{\tilde{G}}$ & pre-trained DeepONet\\
		$\Omega$ & domain of the PDE\\
		$\{x_1, x_2, \cdots, x_m\}$ & scattered sensors\\
		$[v(x_1), v(x_2), \cdots, v(x_m)]$ & input of branch network\\
		$[b_1(v), b_2(v), \cdots, b_p(v)]$ & output of branch network\\
		$\xi$ & input of trunk network\\
		$[t_1(\xi), t_2(\xi),\cdots, t_p(\xi)]$ & outputs of trunk network, where $p$ is the number of neurons \\
% 		$v$ & input functions\\
% 		$u$ & output function\\
		$\tilde{u}$ & prediction using pre-trained DeepONet\\
		$W_2$ & 2-Wasserstein distance\\
		$\mathcal{F}$ & governing PDEs and/or physical constraints\\
		$\mathcal{B}$ & initial and boundary conditions\\
		$\mathcal{D}$ & sparse observations\\
		$\mathcal{E}_{\text{phys}}$ & mismatch error of physics \\
		$\mathcal{E}_{\text{obs}}$ & mismatch error of observations \\
		$\mathcal{L}_{\text{phys}}$ & loss for FT-Phys\\
		$\mathcal{L}_{\mathcal{F}}$ & loss of PDE residuals\\
		$\mathcal{L}_{\mathcal{B}}$ & loss of initial and boundary conditions\\
		$\mathcal{L}_{\text{obs}}$ & loss for FT-Obs-A\\
		$\mathcal{L}_{\mathcal{F},\text{obs}}$ & loss for FT-Obs-T\\
		$w_{\mathcal{F}},\ w_{\mathcal{B}}$ & weights in FT-Phys loss\\
		$\lambda$ & weight in FT-Obs-T loss \\
		\bottomrule
    \end{tabular}
    \label{tab:notation}
\end{table}

\section{Hyperparameters}
\label{sec:hyperparameter}

Table~\ref{table:hyperpara} provides the DeepONet architectures used in all examples and the hyperparameters for training.

\begin{table}[htbp]
\centering
\caption{\textbf{DeepONet architectures and the hyperparameters used for pre-training.} In the ``Depth'' and ``Activation'' columns, the first and second subcolumns correspond to the trunk and branch net, respectively. The branch net and trunk net use the same network width.}
\label{table:hyperpara}
\begin{tabular}{l|ccc|cc}
 \toprule
 Problems & Depth & Width & Activation & Learning rate & Iterations \\
 \midrule
 Section~\ref{subsec:ode} Antiderivative & 3, 3 & 40 & $\tanh$, ReLU & 0.005 & $5\times 10^{4}$ \\
 Section~\ref{subsec:diffusion-reaction} Diffusion-reaction & 4, 3 & 100 & GELU, ReLU & 0.001 & $5\times 10^{5}$ \\
 Section~\ref{subsec:burgers} Burgers' & 4, 3 & 100 & GELU, ReLU & 0.001 & $5\times 10^{5}$ \\
 Section~\ref{subsec:advection} Advection & 4, 3 & 100& GELU, ReLU & 0.001 & $5\times 10^{5}$ \\
 Section~\ref{subsec:poisson} Poisson (Notch) & 4, 3 & 100& GELU, ReLU & 0.001 & $5\times 10^{5}$ \\
 Section~\ref{subsec:lid} Lid-driven cavity & 4, 3 & 100 & GELU, ReLU & 0.001 & $5\times 10^{5}$ \\
 \bottomrule
\end{tabular}
\end{table}

For the method of fine-tuning with physics, we use the Adam optimizer and the number of iterations is listed in Table~\ref{tab:hyper2}. For FT-Obs-A, we fine-tune the DeepONet for 500 iterations using the L-BFGS optimizer for all the problems, except that for the antiderivative problem, we use the Adam optimizer with a learning rate of 0.001 for 1000 iterations. For FT-Obs-T, we choose $\lambda=0.3$ for all cases, and we fine-tune the DeepONet for 3000 iterations using the Adam optimizer with the learning rate of 0.001, except that for the antiderivative problem, we train for 1000 iterations. 

\begin{table}[htbp]
    \caption{\textbf{Hyperparameters for FT-Phys.}}
    \label{tab:hyper2}
    \centering
    \begin{tabular}{l|c}
        \toprule
         & FT-Phys iterations \\
        \midrule
        Section~\ref{subsec:ode} Antiderivative &1000 \\
        Section~\ref{subsec:diffusion-reaction} Diffusion-reaction & 2000\\
        Section~\ref{subsec:burgers} Burgers' & 5000\\
        Section~\ref{subsec:advection} Advection & 5000 \\
        \bottomrule
    \end{tabular}
\end{table}

For multifidelity learning, the size of the low-fidelity dataset $\mathcal{S}$ is in Table~\ref{tab:hyper3}. However, MFGPR is not able to handle a large dataset, and thus we use at most 400 low-fidelity data points. For MFNN, we use the SiLU activation function, and the network size is in Table~\ref{tab:hyper3}. We train MFNN using the Adam optimizer for 10000 iterations. Also, a $L^2$ regularization is applied, and the strength is $10^{-6}$ for the antiderivative operator. For other problems, the strength is $10^{-5}$, $10^{-6}$, $10^{-7}$, and $10^{-8}$ for 20, 50, 100, and 200 high-fidelity data points, respectively. 

\begin{table}[htbp]
    \centering
    \caption{\textbf{Hyperparameters for MFNN.} In the columns of low- and high-fidelity networks, the first and second numbers are depth and width, respectively.}
    \label{tab:hyper3}
    \begin{tabular}{l|cccc}
        \toprule
         & $|\mathcal{S}|$ & Low-fidelity  & High-fidelity & Learning \\
         & &network &network & rate\\
        \midrule
         Section~\ref{subsec:ode} Antiderivative & 100& 4, 40 & 3, 30& 0.005 \\
         Section~\ref{subsec:diffusion-reaction} Diffusion-reaction &10201&4, 128&3, 15& 0.001 \\
        Section~\ref{subsec:burgers} Burgers' &10201&4, 128&3, 15& 0.001\\
        Section~\ref{subsec:advection} Advection &10201&4, 128&3, 15& 0.001 \\
        Section~\ref{subsec:poisson} Poisson (Notch) &5082 &4, 128&3, 15& 0.001 \\
         Section~\ref{subsec:lid} Lid-driven cavity &10201 &4, 128&3, 15& 0.001\\
        \bottomrule
    \end{tabular}
\end{table}

\section{Layer-wise locally adaptive activation function}

\begin{figure}[htbp]
    \centering
    \includegraphics[width=\textwidth]{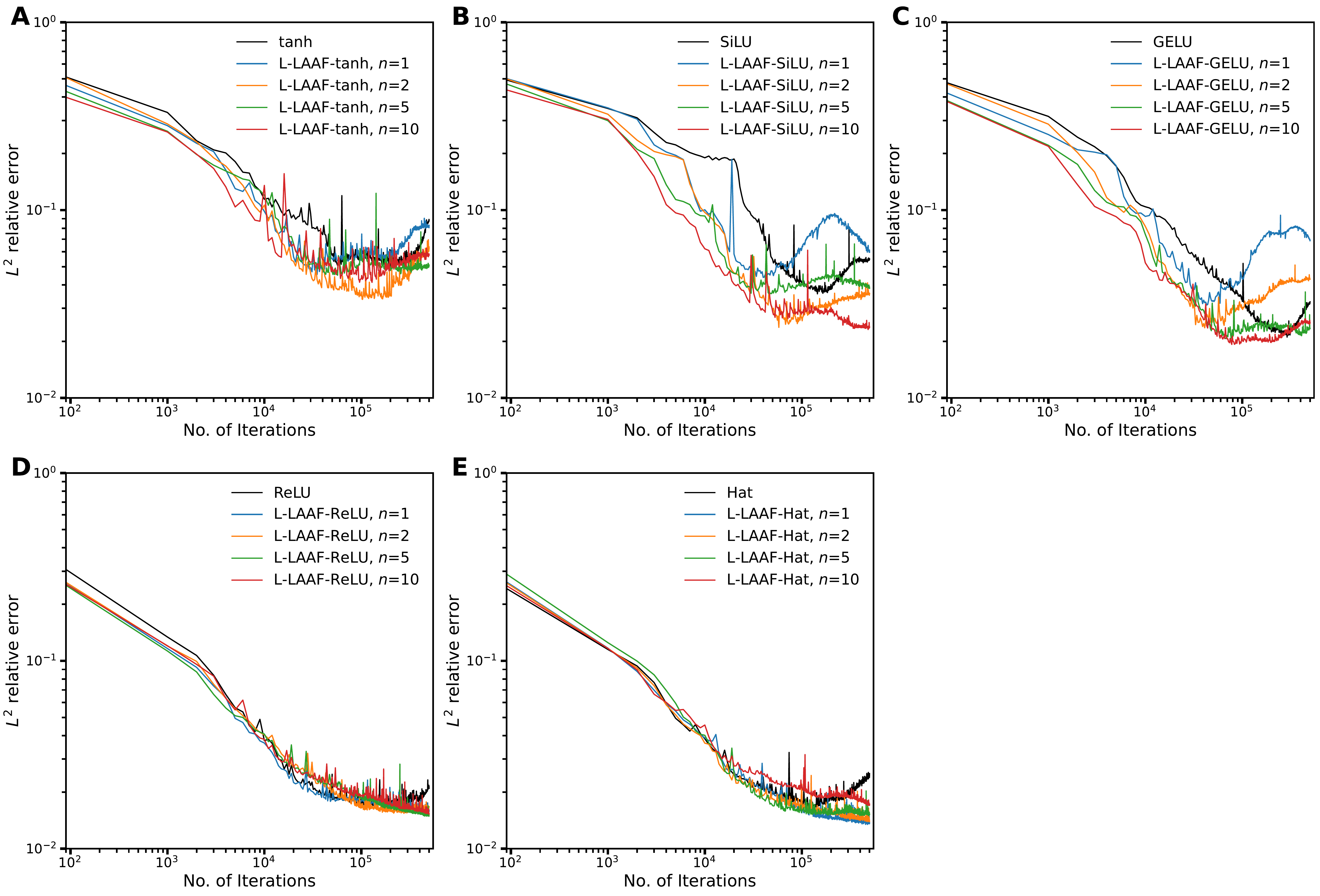}
     \caption{\textbf{Section~\ref{sec:activation}: L-LAAF with different scaling factors $n$.} (\textbf{A}) $\tanh$. (\textbf{B}) SiLU. (\textbf{C}) GELU. (\textbf{D}) ReLU. (\textbf{E}) Hat.}
     \label{fig:L-LAAF}
\end{figure}

For L-LAAF in Section~\ref{sec:activation}, the scaling factor $n$ is a hyperparameter to be tuned. We choose $n$ from 1, 2, 5, and 10 for each activation function. The best scaling factors $n$ for $\tanh$, SilU, GELU, ReLU, and Hat are 2, 10, 10, 5, and 1, respectively (Fig.~\ref{fig:L-LAAF}).

\section{Fine-tune with physics}
\label{sec: ft-phys detail}

\begin{figure}[htbp]
    \centering
    \includegraphics[width=\textwidth]{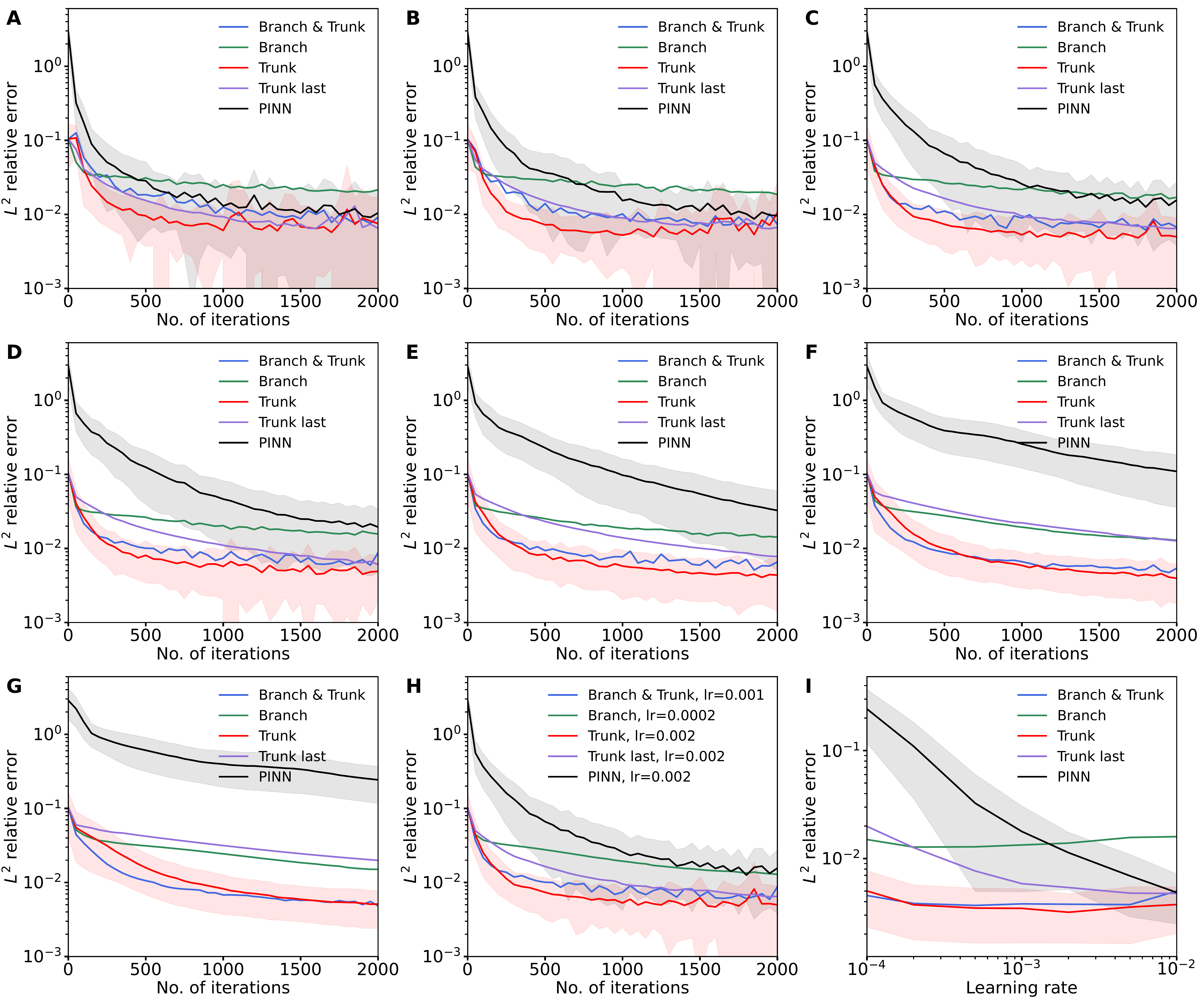}
     \caption{\textbf{Section~\ref{subsec:diffusion-reaction}: Fine-tuning with physics for the diffusion-reaction equation.}  (\textbf{A--G}) Training trajectories under different learning rate of (A) 0.01, (B) 0.005, (C) 0.002, (D) 0.001, (E) 0.0005, (F) 0.0002, and (G) 0.0001. (\textbf{H}) The best result of each method among different learning rates. (\textbf{I}) $L^2$ relative errors with respect to learning rate for fine-tuning different parts of DeepONet. The curves and shaded regions represent the mean and one standard deviation of 100 runs. For clarity, only standard deviations of trunk mode and PINN mode are plotted.}
     \label{fig:diff-reac}
\end{figure}

\begin{figure}[htbp]
    \centering
    \includegraphics[width=\textwidth]{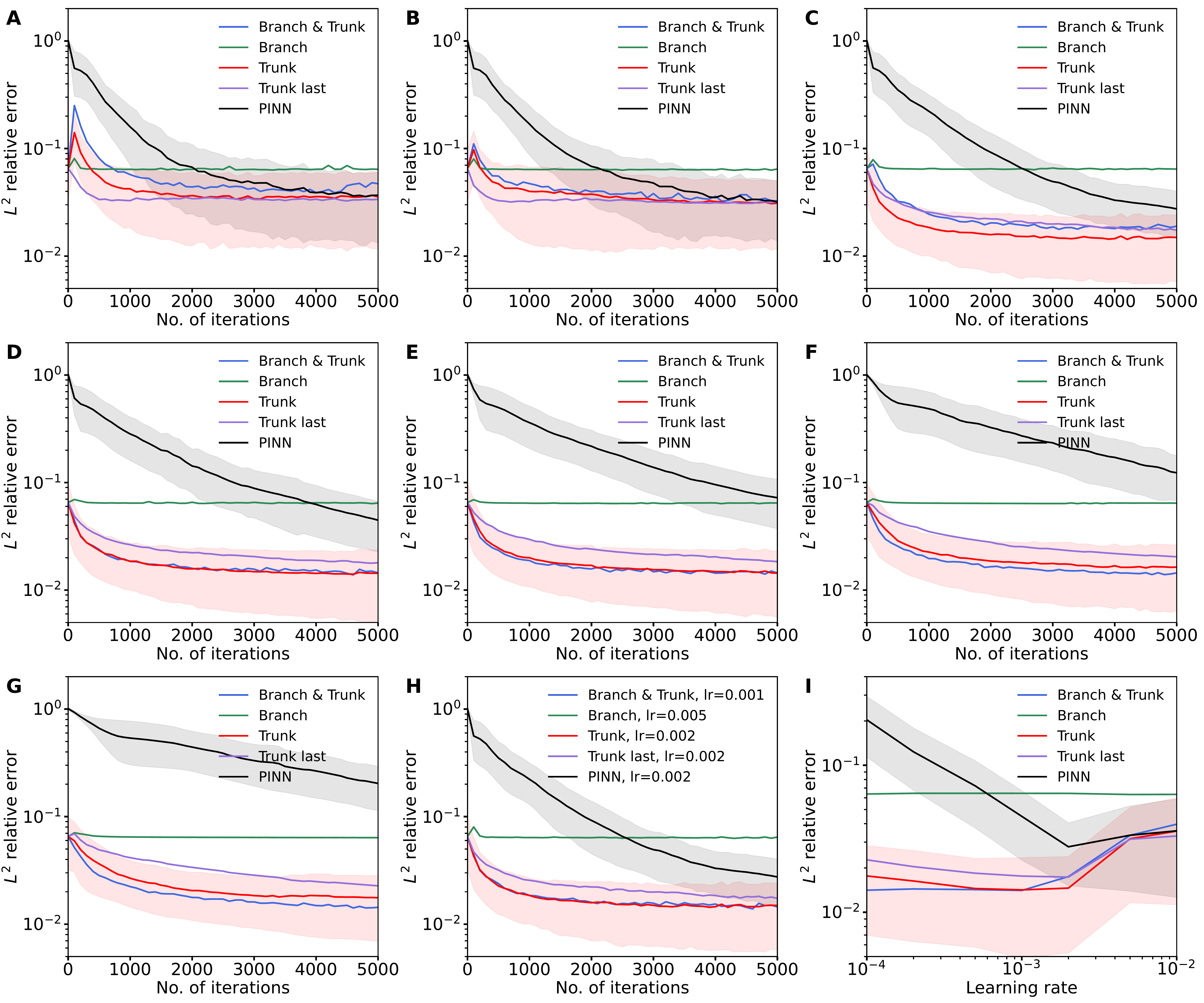}
    \caption{\textbf{Section~\ref{subsec:burgers}: Fine-tuning with physics for the Burgers' equation.}  (\textbf{A--G}) Training trajectories under different learning rate of (A) 0.01, (B) 0.005, (C) 0.002, (D) 0.001, (E) 0.0005, (F) 0.0002, and (G) 0.0001. (\textbf{H}) The best result of each method among different learning rates. (\textbf{I}) $L^2$ relative errors with respect to learning rate for fine-tuning different parts of DeepONet. The curves and shaded regions represent the mean and one standard deviation of 100 runs. For clarity, only standard deviations of trunk mode and PINN mode are plotted.}
    \label{fig:burgers}
\end{figure}

\begin{figure}[htbp]
    \centering
    \includegraphics[width=\textwidth]{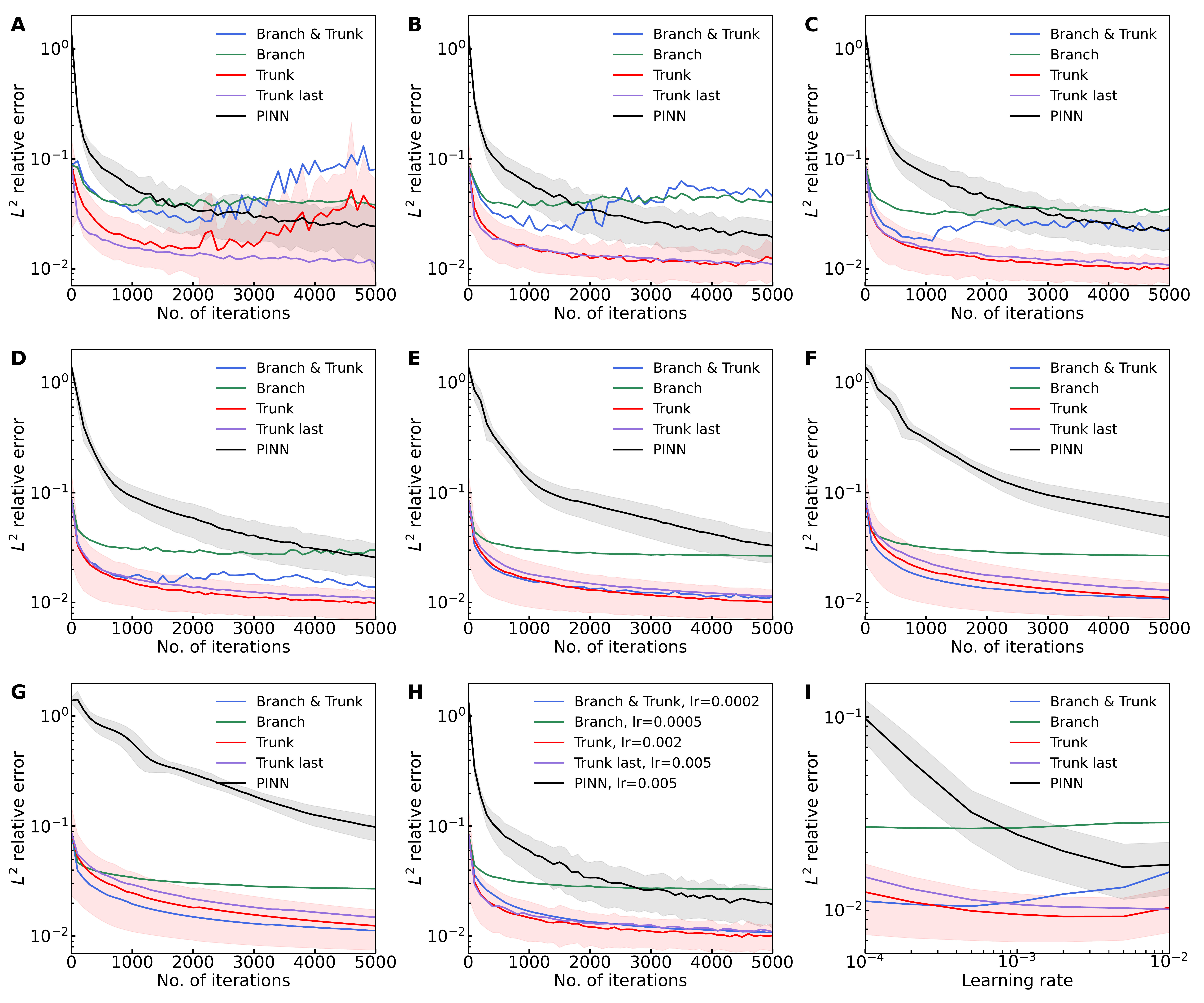}
    \caption{\textbf{Section~\ref{subsec:advection}: Fine-tuning with physics for the advection equation.}  (\textbf{A--G}) Training trajectories under different learning rate of (A) 0.01, (B) 0.005, (C) 0.002, (D) 0.001, (E) 0.0005, (F) 0.0002, and (G) 0.0001. (\textbf{H}) The best result of each method among different learning rates. (\textbf{I}) $L^2$ relative errors with respect to learning rate for fine-tuning different parts of DeepONet. The curves and shaded regions represent the mean and one standard deviation of 100 runs. For clarity, only standard deviations of trunk mode and PINN mode are plotted.}
    \label{fig:advection}
\end{figure}

With physics as additional information, fine-tuning with physics is considered and different learning rates (i.e., 0.01, 0.005, 0.002, 0.001, 0.0005, 0.0002, and 0.0001) are used to fine-tune the pre-trained DeepONet. Below are detailed results of diffusion reaction equation (Fig.~\ref{fig:diff-reac}), Burgers' equation (Fig.~\ref{fig:burgers}), and advection equation (Fig.~\ref{fig:advection}).

\section{Comparisons of different values of $\lambda$ for FT-Obs-T methods}
\label{sec:dr_lambda}

\begin{figure}[htbp]
    \centering
    \includegraphics[width=\textwidth]{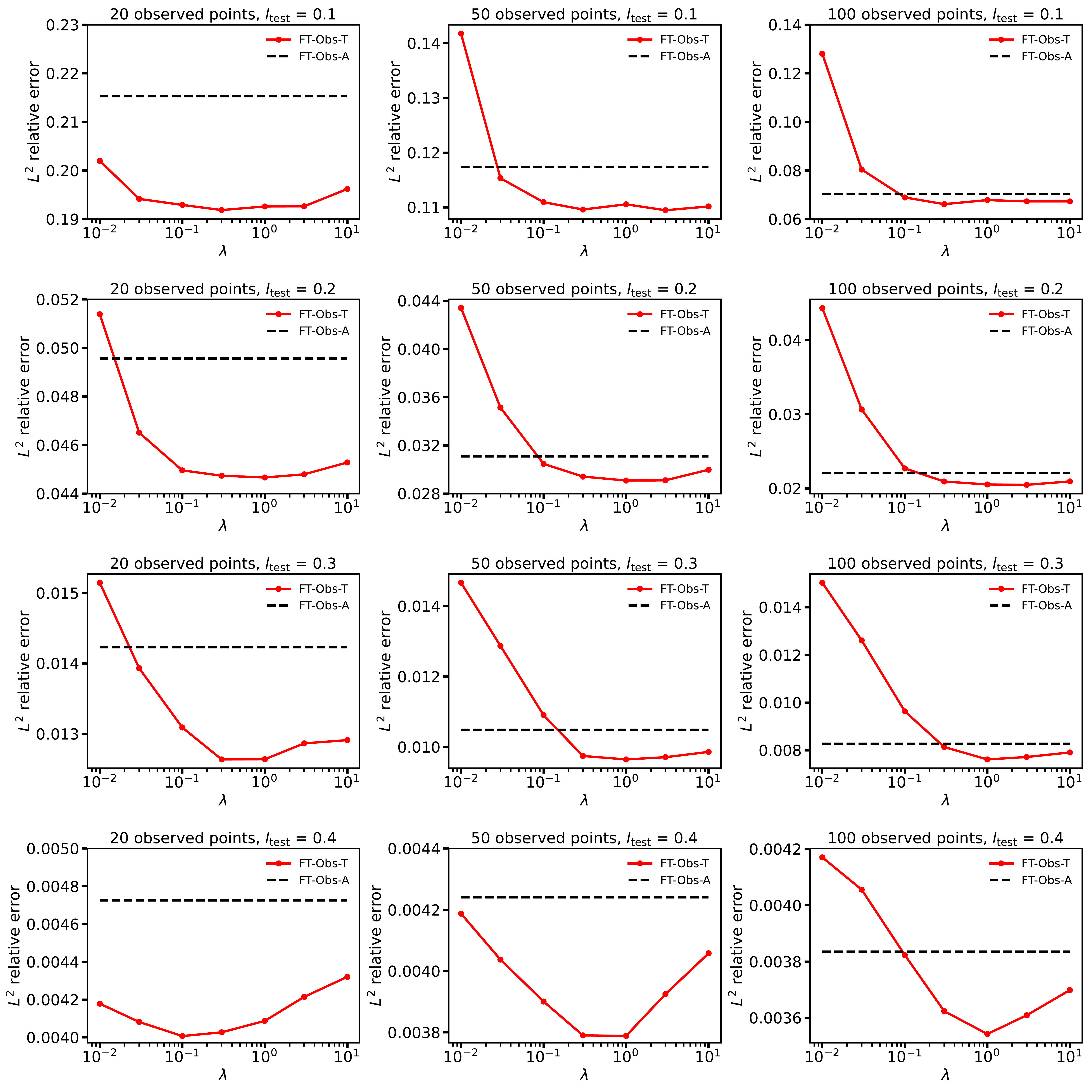}
        \caption{\textbf{Section~\ref{subsec:diffusion-reaction}: Comparisons of different values of $\lambda$ for FT-Obs-T method under different number of observed points and testing correlation lengths for the diffusion-reaction equation.} Different rows represent different testing correlation lengths. Different columns represent different numbers of new observations.}
    \label{fig:k}
\end{figure}

For diffusion-reaction equation in Section~\ref{subsec:diffusion-reaction}, we further determine the effect of $\lambda$ on test errors when using different numbers of observed points and different $l_{\text{test}}$. Results of 12 cases are shown in Fig.~\ref{fig:k}.

\section{MFGPR for the Lid-driven cavity flow}
\label{sec:kernel_MFGPR}

\begin{figure}[htbp]
    \centering
    \includegraphics[width=.8\textwidth]{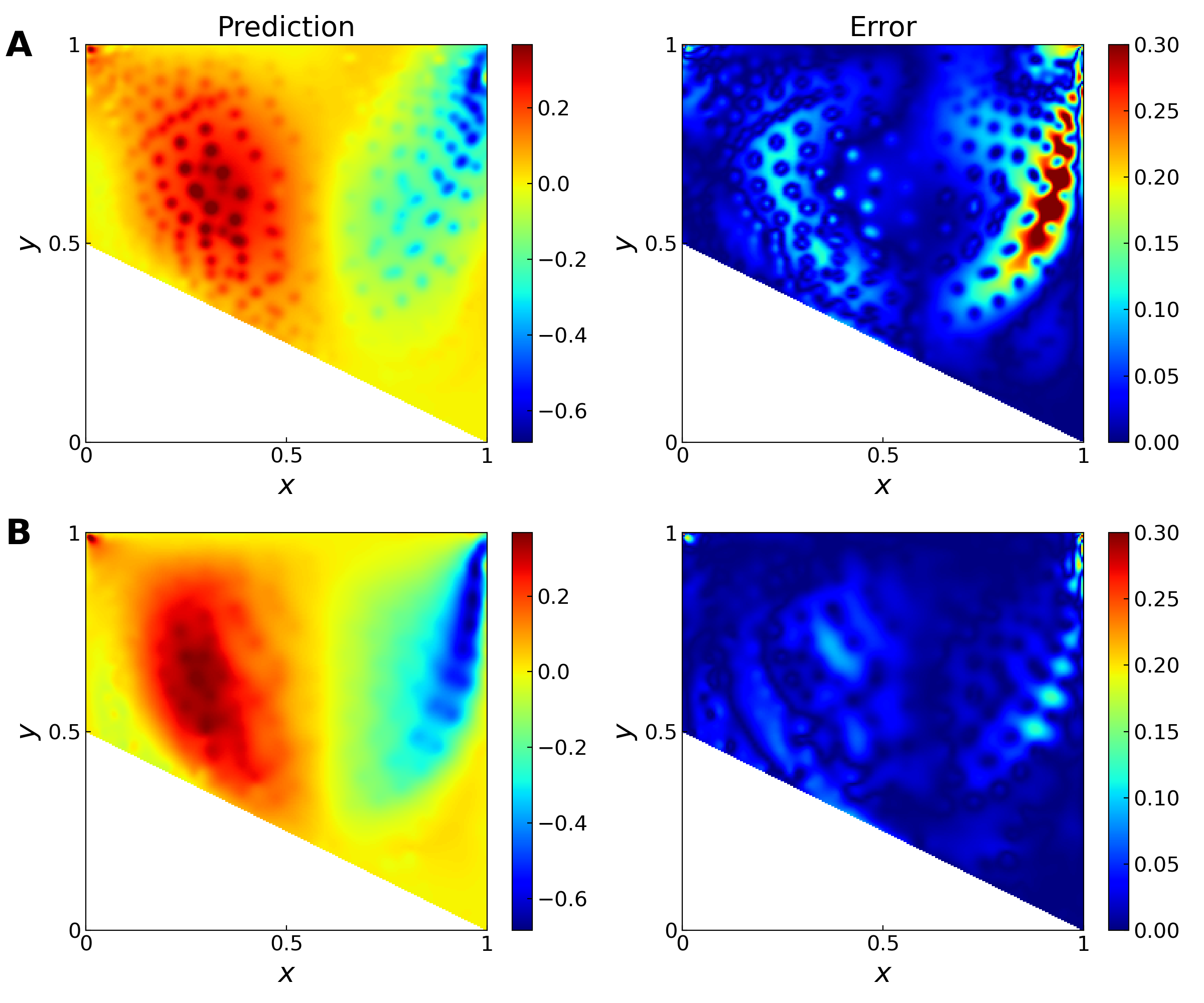}
    \caption{\textbf{Section~\ref{subsec:lid}: Prediction and error of MFGPR with the RBF and Matern kernels for the lid-driven cavity flow.} (\textbf{A}) The RBF kernel. (\textbf{B}) The Matern kernel with $\nu=1.5$.}
    \label{fig:cavity_mfgp}
\end{figure}

In most examples, the RBF kernel works well for MFGPR. However, in the cavity flow problem, MFGPR with the RBF kernel has a large error (Fig.~\ref{fig:cavity_mfgp}A), while the Matern kernel with $\nu=1.5$ performs better (Fig.~\ref{fig:cavity_mfgp}B). The Matern kernel is given by
$$k(x_1, x_2) = \frac{1}{\Gamma(\nu)2^{\nu-1}} \left(\frac{\sqrt{2\nu}}{l} \|x_1 - x_2\| \right)^\nu K_\nu \left(\frac{\sqrt{2\nu}}{l} \|x_1 - x_2\| \right),$$
where $l$ is the correlation length, $K_{\nu}(\cdot)$ is a modified Bessel function, and $\Gamma(\cdot)$ is the gamma function.

\bibliographystyle{unsrt} 
\bibliography{main}

\end{document}